\documentclass{article}
\usepackage{utils/template}
\usepackage{times}
\usepackage{latexsym}
\usepackage[T1]{fontenc}
\usepackage[utf8]{inputenc}
\usepackage{microtype}
\usepackage{url}
\usepackage{inconsolata}
\usepackage{comment}
\usepackage{subcaption}
\usepackage{graphicx}
\usepackage{color, colortbl}
\usepackage{multirow}
\usepackage{wrapfig}
\usepackage{enumitem}

\newcommand{\texttoimagemodel}{Imagen}
\newcommand{\crowdsource}{CrowdSourced}
\newcommand{\waa}{Webli-Align}
\newcommand{\pd}{Prefix Dropout}
\newcommand{\rft}{Re-fine-tuned}

\definecolor{cvprblue}{rgb}{0.21,0.49,0.74}
\usepackage[breaklinks,colorlinks,citecolor=cvprblue]{hyperref}

\definecolor{ci1}{rgb}{0.07 , 0.33 , 0.8 }
\definecolor{ci2}{rgb}{0.8 , 0. , 0. }
\definecolor{ci3}{rgb}{0.9 , 0.57 , 0.22 }
\definecolor{ci4}{rgb}{0.56 , 0.49 , 0.76 }
\definecolor{ci5}{rgb}{0.65 , 0.30 , 0.47 }
\definecolor{ci6a}{rgb}{0.27 , 0.51 , 0.56 }
\definecolor{ci6b}{rgb}{0.04 , 0.33 , 0.58 }
\definecolor{ci6c}{rgb}{0.65 , 0.11 , 0. }

\title{Prompt Expansion for \\ Adaptive Text-to-Image Generation}

\usepackage[bottom]{footmisc}
\usepackage{authblk}

\author[ ]{\bf Siddhartha Datta$^{\tiny \textnormal{2}}$\thanks{Work done while at Google. Correspondence with author at \url{siddhartha.datta@cs.ox.ac.uk}.}}
\author[1, 3]{\bf Alexander Ku}
\author[1]{\bf Deepak Ramachandran}
\author[1]{\bf Peter Anderson}
\affil[1]{Google Research}
\affil[2]{University of Oxford}
\affil[3]{Princeton University}

\begin{document}
\begin{figure*}[!b]
    \hspace{-2.75cm}\includegraphics[width=1.45\textwidth]{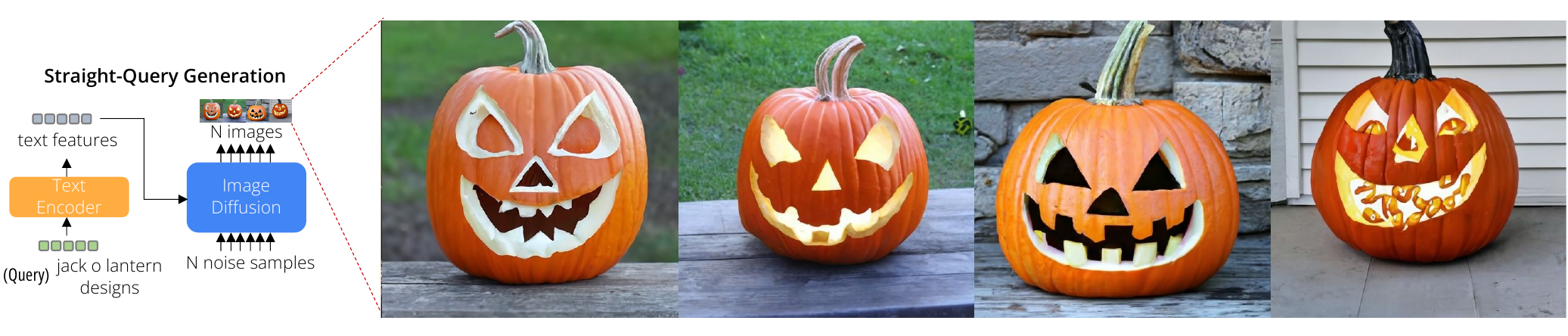}
    
    \hspace{-2.75cm}\includegraphics[width=1.45\textwidth]{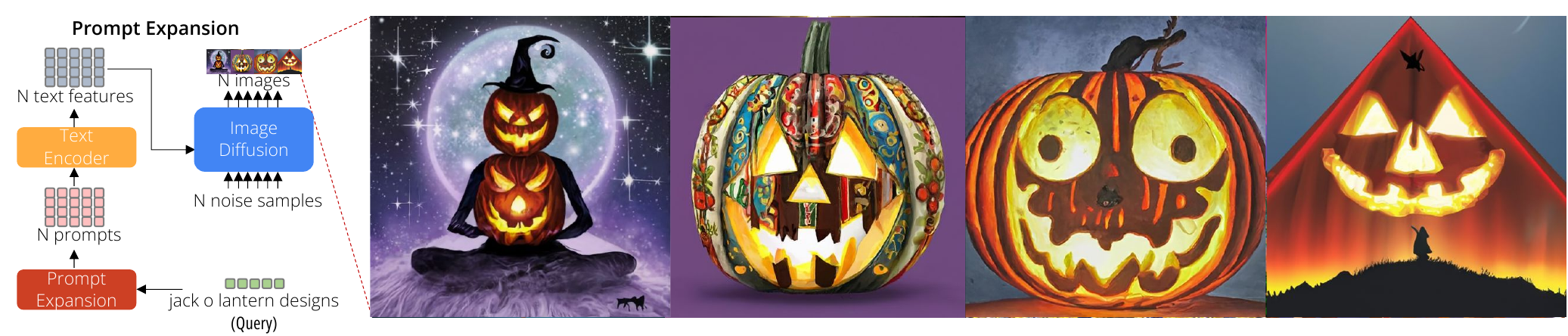}
\caption{
Prompt Expansion is an alternative paradigm for interaction with text-to-image models. 
\textbf{Top}: Sampled outputs from Straight-Query Generation may result in images that are less visually compelling and diverse. \textbf{Bottom}: Prompt Expansion samples uncommitted aspects of the image in text space, improving visual quality and diversity while enabling interaction modes beyond prompt engineering / iteration.
}
\label{fig:concept}
\end{figure*}

\maketitle
\begin{abstract}
Text-to-image generation models are powerful but difficult to use. Users craft specific prompts to get better images, though the images can be repetitive. This paper proposes a Prompt Expansion framework that helps users generate high-quality, diverse images with less effort. The Prompt Expansion model takes a text query as input and outputs a set of expanded text prompts that are optimized such that when passed to a text-to-image model, generates a wider variety of appealing images.
We conduct a human evaluation study that shows that images generated through Prompt Expansion are more aesthetically pleasing and diverse than those generated by baseline methods. Overall, this paper presents a novel and effective approach to improving the text-to-image generation experience.
\end{abstract}

\clearpage
\clearpage
\begin{figure*}[t]
    \centering
    \hspace*{-4.25cm}
    \includegraphics[width=1.6\textwidth]{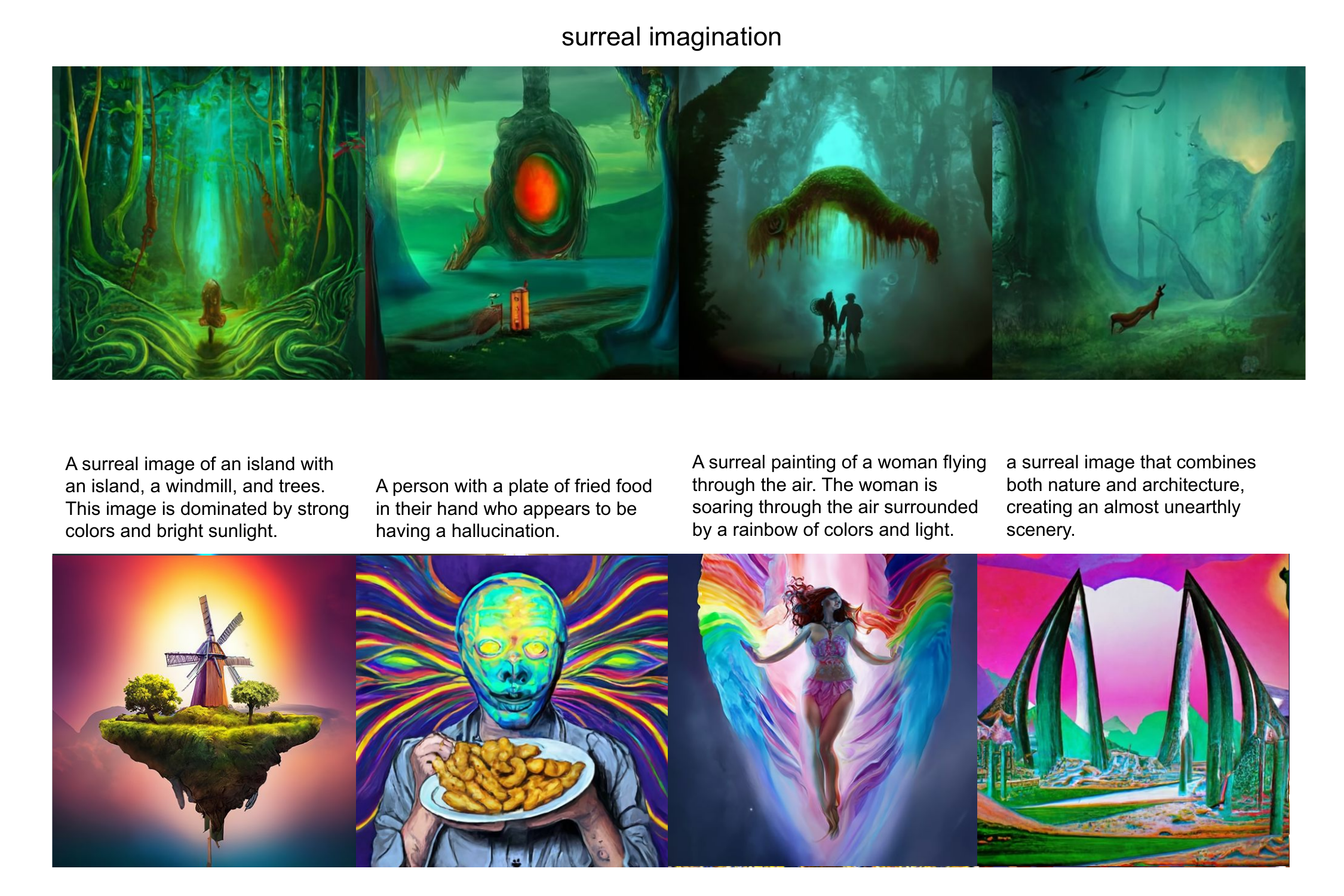}
    \hspace*{-3.3cm}
    \includegraphics[width=1.47\textwidth]{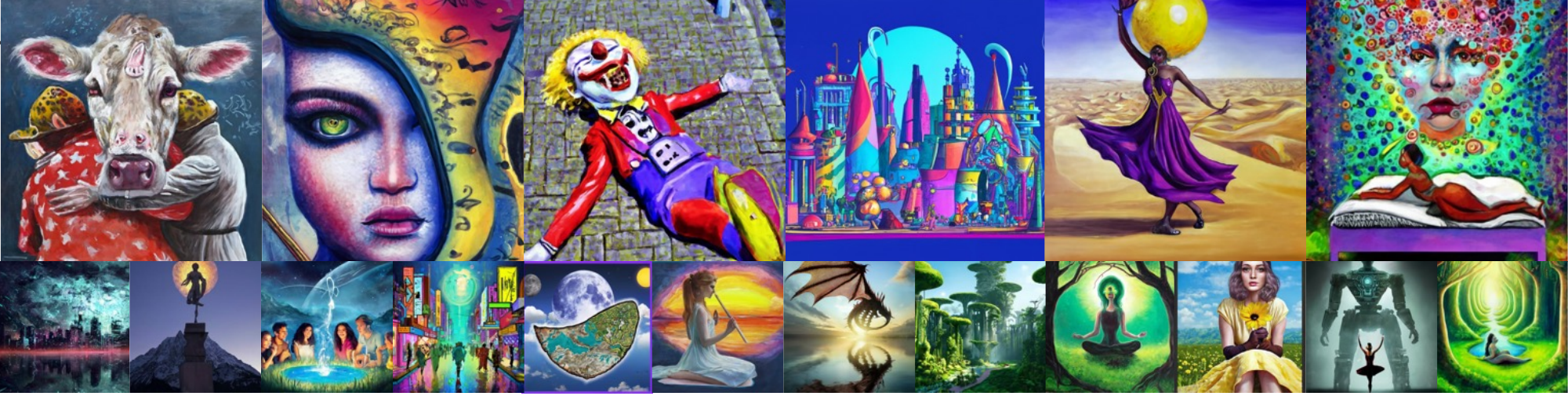}
    \captionsetup{width=1.5\textwidth}
    \caption{
    \textbf{Abstract queries}:
    While Straight-Query Generation returns a set of similar images (Top),
    Prompt Expansion returns a diverse range of images (Bottom), varying in topic, subject, location, colours, etc.
    This range is reflected in the expanded prompts for the first four images shown to the user, 
    and is continued to be reflected in the next 18 images generated.
    }
    \label{fig:demo1}
\end{figure*}

\begin{figure*}[t]
    \centering
    \vspace*{-3.2cm}
    \captionsetup{width=1.5\textwidth}
    \hspace*{-2cm}
    \includegraphics[width=1.25\textwidth]{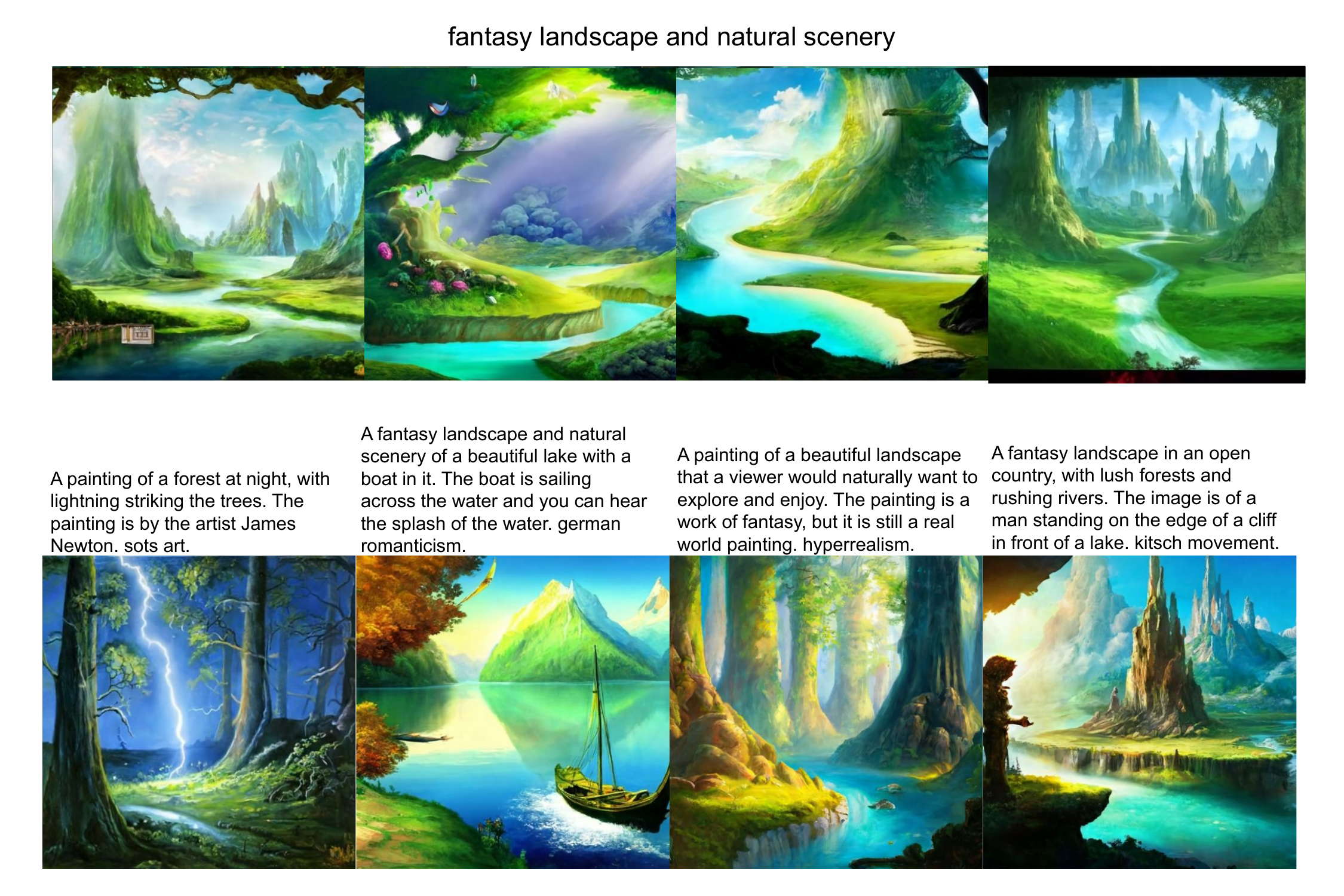}
    \vspace*{-0.75cm}
    \caption{
    \textbf{Abstract-Concrete queries}:
    For a query that is broad in topic but contains specific details that should not be left out, 
    straight-query generation still returns similar-looking images.
    Prompt Expansion returns images that remain entailed and faithful to the query, but adds variations in lighting, fine-grained details. 
    }
    \label{fig:demo2}
\end{figure*}

\begin{figure*}[t]
    \centering
    \vspace*{-0.75cm}
    \hspace*{-2cm}
    \includegraphics[width=1.25\textwidth]{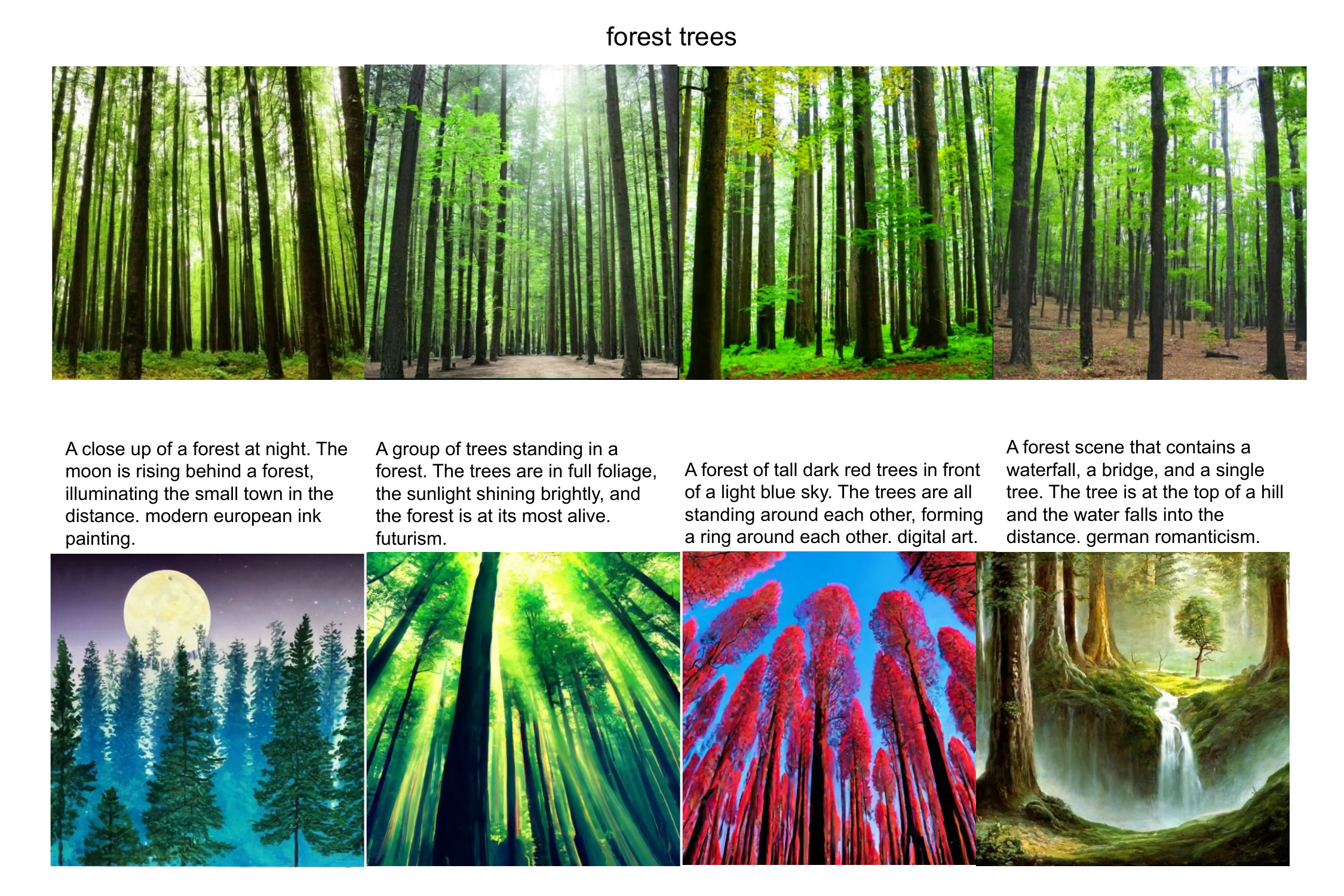}
    \captionsetup{width=1.5\textwidth}
    \vspace*{-0.75cm}
    \caption{
    \textbf{Concrete queries}:
    For a query that is very specific in nature, 
    Straight-Query Generation (Top) returns images of the same forest setting with trees from the same angle, same species, ground vegetation, etc.
    Prompt Expansion (Bottom) explores colours, styles of imagery, angles of view, time of day, scenery, etc.
    }
    \label{fig:demo3}
\end{figure*}

\clearpage

\section{Introduction}
\label{sec:intro}

Text-to-image generation models~\citep{ramesh-dalle2, imagen-paper, yu-parti} are capable of rendering a stunning variety of high-quality images, from highly-realistic professional-looking photos to fanciful dreamworlds in almost any visual style. However, interaction with these models frequently exposes two significant usability problems. First, the connection between text inputs and image outputs in these models still bears the fingerprints of the training data, which is typically images scraped from the web along with their alt-text. As a consequence, achieving high-quality outputs often requires users to include arcane lighting and camera jargon (`35mm', `DSLR', `backlit' `hyper detailed'), idiosyncratic descriptors (`audacious and whimsical'), and social media tags (`trending on artstation'). The prompts that produce the best images are not necessarily stable across different models, or even across different versions of the same model, leading to a focus on prompt-sharing and `prompt engineering', which is the process of iterating over prompts to craft the optimal textual input for a desired image.

Second, we find that when sampling sets of images from these models, while the outputs are \emph{different}, they are not necessarily \emph{diverse}. For example, randomly sampling four images for the prompt `jack o lantern designs' from a diffusion model similar to \texttoimagemodel~\citep{imagen-paper} produces highly similar outputs (refer Fig \ref{fig:concept} Top). Even though the prompt does not specify the composition, style, viewpoint or lighting of the image, whether it is night or day, or whether there are other features in the scene, samples from the model do not vary in any of these aspects. This lack of diversity can amplify harmful social biases when generating people (e.g. generating all doctors as male)~\citep{naik2023social}, and also fails to lead the user towards more interesting imagery. Just as people are non-committal about basic aspects of their mental images \citep{BIGELOW2023105498}, we argue that text-to-image models should also exhibit non-commitment by generating diverse outputs for any image characteristics that are unspecified~\citep{hutchinson-etal-2022-underspecification}.

We reduce the burden of prompt engineering and improve diversity in text-to-image generation by proposing a \emph{Prompt Expansion} (PE) framework. A Prompt Expansion model takes a text prompt as input, which we refer to as a query, and outputs a set of $N$ expanded text prompts that include specialized keywords (to improve image quality) and interesting additional details (to add diversity to the generated images). 
To do this, we construct a Prompt Expansion dataset by inverting a dataset of high aesthetic images to text, and few-shot prompting the inverted text to queries. We then train a PALM 2~\citep{anil2023palm} text-to-text model on the query:prompt pairs, and iteratively re-fine-tune based on generated output.
As illustrated in Fig \ref{fig:concept} Bottom, incorporating Prompt Expansion into text-to-image generation produces a greater variety of appealing, high-quality images and also opens up additional modes of user interaction beyond iterating over prompts (e.g., the user can simply select the image they prefer, which can be used as the starting point for another iteration of Prompt Expansion).
Empirically, we observe that Prompt Expansion achieves improved diversity, aesthetics, and text-image alignment compared to Straight-Query Generation in both automatic metrics and human rater evaluation.

\noindent
\textbf{Contributions.}
Motivated by the frustrations of prompt-sharing and prompt-engineering, we:
\begin{enumerate}
\item Propose a new framework, Prompt Expansion, to improve image quality and diversity while opening up new modes of interaction;
\item Construct a Prompt Expansion dataset by reverse engineering prompts and high-level queries from aesthetic images, and train a variety of Prompt Expansion models; 
\item Undertake a detailed human evaluation demonstrating that Prompt Expansion can improve diversity, aesthetics, and text-image alignment.
\end{enumerate}

\section{Related Work}

To better control and introduce diversity into image output, 
most prior work focuses on techniques to allow end-users to edit the generated images, such as 
text-guided image editing~\citep{hertz2022prompt},
or to fine-tune to create personalized images~\citep{collell-coling-2016, ruiz2022dreambooth}.
These methods focus on local edits and iterating with respect to a given image.
Our goal is to introduce diversity in the generated images set.
Modifying the sampler is a common approach to achieving this.
The standard approaches to introducing variation are 
increasing guidance~\citep{ho2021classifierfree} in diffusion models (e.g. Imagen~\citep{imagen-paper}),
or increasing temperature in autoregressive models (e.g. Parti~\citep{yu-parti}).
Changing the sampler/decoder hyperparameters, however, may not return meaningfully-diverse images.
Setting the temperature/guidance too high will insert excessive noise and yield images of low quality.
The noise inserted is random, thus it may not return variations along targeted dimensions (e.g. style, lighting, attributes of people / objects / places). 
To interpretably sample images, we shift our attention to sampling within text space. 
\begin{wrapfigure}{r}{0.5\textwidth}
    \centering
    \vspace{-10pt}
    \includegraphics[width=0.5\textwidth]{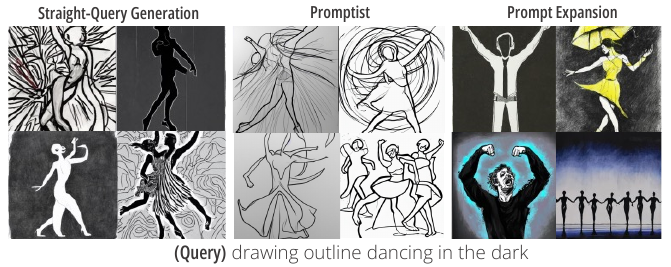}
    \caption{
    We demonstrate the difference between prompt optimization techniques and prompt exploration techniques.
    While prompt optimization methods such as Promptist can return the optimized prompt \textit{"drawing outline dancing in the dark, trending on artstation"},
    it aims to return one optimized prompt, and thus 4 images generated from this optimized prompt may return less visual diversity and explores less of the prompt space than Prompt Expansion.
    We generate Promptist's optimized prompt with Stable Diffusion 1.4, which is the model it is aligned with during its training.
}
\vspace{-10pt}
    \label{fig:promptist}
\end{wrapfigure}
Prompt Optimization techniques such as
Promptist~\citep{hao2022optimizing}
or Promptify~\citep{brade2023promptify}
investigate automatic and human-in-the-loop approaches to improving the aesthetic quality of a prompt's generated image.
The task of Prompt Optimization
presumes that an ideal prompt may exist for a given query that returns the "best" image.
Instead, the task of Prompt Expansion 
acknowledges that the ideal generated image depends on more than the query alone and that the query may have different intents or context (e.g. varying by user preferences). It addresses the fact that there are
uncommitted aspects of the image (e.g. query ambiguity) and that images are evaluated in sets.
It also lays a foundation for treating image generation as a multi-turn process.As an example, 
an abstract query "drawing outline dancing in the dark" in
Figure \ref{fig:promptist}
returns a similar set of black-and-white dancers.
Prompt Optimization (e.g. Promptist) would return a prompt that maximizes aesthetics, but it would still return a similar set of black-and-white dancers.
PE explores 
variations and possibilities 
in colors, angles, 
level of abstraction, gender, plurality, etc.

\section{Prompt Expansion Dataset}
\label{sec:pemdataset}

The \textit{Prompt Expansion} (PE) framework requires a model to take a user text \textit{query} as input and return $N$ text \textit{prompts} as output, such that the $N$ text prompts through text-to-image generation will return a set of $N$ diverse, aesthetic images aligned to the query.
To train a PE model,
we require a dataset mapping queries to prompts, which we construct in reverse.
First, we collect aesthetically high-quality images, including both model-generated and natural images (refer Sec \ref{sec:aesthetic}).
Second, we invert the images to a closely corresponding prompt that includes alt-text jargon (which we refer to as \textit{flavors}, refer Sec \ref{sec:inversion}). 
Finally, we map the inverted text to a range of high-level queries that more closely correspond to user input (refer Sec \ref{sec:extraction}). These queries are paired with the prompts from the second step to form the \texttt{\{query:prompt\}} dataset.

\subsection{Image-Aesthetic Datasets}
\label{sec:aesthetic}

We curate two image datasets.
The first, \waa{}, 
is composed of images from the Webli~\citep{chen2023pali} and Align~\citep{jia2021scaling} datasets, filtered to retain only images with high MUSIQ~\citep{ke2021musiq} aesthetic scores.
The second, \crowdsource{}, is obtained by crowd-sourcing output from a text-to-image model. We provide an interface for text-to-image generation similar to \citet{gradio}, allowing users from a large organization to enter prompts that generate images. Users also have the option to upvote images that they like. We use this signal to retain only the most appealing images.
We retain 80k \waa{} (natural)
and 40k \crowdsource{} (generated)
images.

\subsection{Image-to-Text Inversion}
\label{sec:inversion}

The second step is to invert the images in the Image-Aesthetic datasets
to prompt text.
While the user \textit{query} is the input a user provides,
the \textit{prompt} is the text that generates a specific image.
We use the \textit{Interrogator} \citep{clipin} approach to image-to-text inversion.
The computed prompt text is generated by concatenating: (i) a caption, and (ii) a set of `flavors'.
The caption is a description of the content of the image (e.g. who, what, where, when).
To generate the caption, we use COCA~\citep{yu2022coca} fine-tuned for the captioning task.
A \textit{flavor} refers to a descriptive word/phrase that alters the style of the image, without intending to add/change the content of the image, like "impressionism" or "dslr".
We generate the lists of flavors from words and phrases used in a large number of collected prompts of generated images (details in Section \ref{sec:interog}).

\begin{table*}[t]
\caption{
Types of prefixes used for controllable generation. The models used in the paper make use of \texttt{ABST}, \texttt{DTL} (* indicates we replace the non-parenthesed prefix with the paranthesed prefix), \texttt{MSTP}.
}
\label{tab:prefixes}
    \begin{subtable}[h]{\textwidth}
        \centering
        \resizebox{\textwidth}{!}{%
        \begin{tabular}{clp{12.5cm}}
        \hline \hline
          \textbf{Prefix}
          & \textbf{Full name} 
          & \textbf{Description} \\ \hline
\texttt{}
{\color{purple} \texttt{ABST}}
& \underline{ABST}ract
& Returns output sentences conditioned on the input sentence being an abstract query (output sentence structure is in-line with the abstract augmentation/mixture in training).
\\
{\color{purple} \texttt{DTL}}
& \underline{D}e\underline{T}ai\underline{L}ed
& Adding this prefix returns output sentences conditioned on the input sentence being a detailed query and requiring a detailed expansion (output sentence structure is in-line with the detailed augmentation/mixture in training).
\\
{\color{purple} \texttt{GRD}} (\texttt{DTL}*)
& \underline{GR}oun\underline{D}ed
& Adding this prefix returns output sentences conditioned on the input sentence being a query requiring a grounded expansion (output sentence structure is in-line with the grounded augmentation/mixture in training).
\\
{\color{purple} \texttt{SPCT}} (\texttt{DTL}*)
& \underline{SP}e\underline{C}ifici\underline{T}y
& Adding this prefix returns output sentences conditioned on the input sentence being a query requiring an expansion that elicits specificity (output sentence structure is in-line with the specificity augmentation/mixture in training).
\\
{\color{purple} \texttt{FLV}}
& \underline{FL}a\underline{V}or
& Adding this prefix returns only the flavor alone.
\\
{\color{purple} \texttt{HAST}}
& \underline{H}igh\underline{A}e\underline{ST}hetics
& Adding this prefix returns an output sentence that should return images with good aesthetics (MUSIC > 6).
\\
{\color{purple} \texttt{RFT}}
& \underline{R}e-\underline{F}ine-\underline{T}uned
& Adding this prefix in front of the the input sentence will return images based on the re-fine-tuning objectives (e.g. based on aesthetics and renderable flavors).
\\
{\color{purple} \texttt{MSTP}}
& \underline{M}ulti\underline{ST}e\underline{P}
& Adding this prefix returns an output sentence specifically for multi-step Prompt Expansion.
\\
\hline \hline
        \end{tabular}%
        }
    \end{subtable}
\end{table*}
\begin{table*}[t]
\caption{
Examples of prefixed \texttt{\{query:prompt\}} pairs, where prepending the {\color{purple} prefix} controls the output generated.
}
\label{tab:examples_augmentations}
    \begin{subtable}[h]{\textwidth}
        \centering
        \resizebox{\textwidth}{!}{%
        \begin{tabular}{p{1.8cm}|p{7cm}p{7cm}}
        \hline \hline
          \textbf{Augmentation}
          & \textbf{Query} 
          & \textbf{Prompt} \\ \hline

Abstract
& \textit{{\color{purple} ABST} hope} & \textit{a tunnel with a light at the far end}
\\\hline \hline
Grounded
& \textit{{\color{purple} GRD} colossal dragon} & \textit{a colossal dragon eating humans}
\\
& \textit{{\color{purple} GRD} monster bike} & \textit{a monster bike being ridden by a daredevil}
\\\hline \hline
Specificity
& \textit{{\color{purple} SPCT} animal drawings} & \textit{animal drawings, for example a children's drawing of a dog}
\\
& \textit{{\color{purple} SPCT} groups of animals} & \textit{groups of animals, specifically a herd of sheep being gathered by a farmer}
\\\hline \hline
Flavor
& \textit{{\color{purple} FLV} a drawing of a dog swimming in a river} & \textit{pointillism}
\\
& \textit{{\color{purple} FLV} a giant robot fighting a dinosaur} & \textit{pixel art}
\\\hline \hline
Multi-step
& \textit{{\color{purple} MSTP} a brain on a wall} & \textit{a brain on a wall, poster art by Robert Beatty}
\\
expansion & \textit{{\color{purple} MSTP} a brain on a wall, poster art by Robert Beatty} & \textit{a brain on a wall, poster art by Robert Beatty, featured on behance}
\\
& \textit{{\color{purple} MSTP} a brain on a wall, poster art by Robert Beatty, featured on behance} & \textit{a brain on a wall, poster art by Robert Beatty, featured on behance, psychedelic art, psychedelic artwork, brains, 8k archival print}
\\
& \textit{{\color{purple} MSTP} a brain on a wall, poster art by Robert Beatty, featured on behance, psychedelic art, psychedelic artwork, brains, 8k archival print} & \textit{a picture of a brain on a wall, poster art by Robert Beatty, featured on behance, psychedelic art, psychedelic artwork, brains, 8k archival print}
\\\hline \hline
        \end{tabular}%
        }
    \end{subtable}
\end{table*}

\subsection{Query/Prompt Extraction}
\label{sec:extraction}

\begin{figure*}[tp]
    \centering
    \hspace*{-2cm}
    \includegraphics[width=1.3\textwidth]{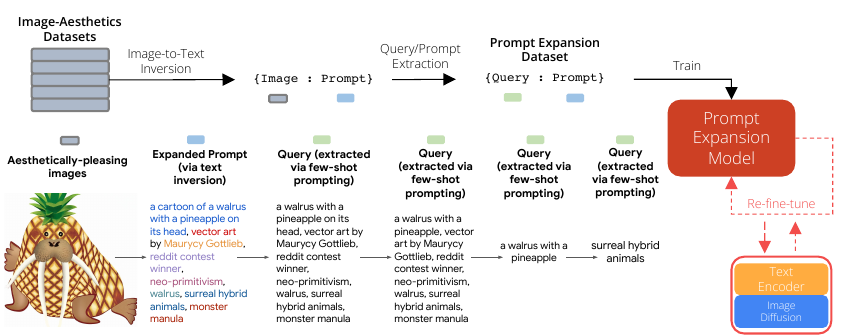}
    \caption{
    \textit{(Outer)}
    Overview of Prompt Expansion dataset construction and model training.
    Beginning with the Image-Aesthetic datasets (Sec \ref{sec:aesthetic}),
    we generate expanded prompts through Image-to-Text Inversion (Sec \ref{sec:inversion}),
    then we extract queries
    for each prompt through Query/Prompt Extraction (Sec \ref{sec:extraction}).
    The resulting Prompt Expansion dataset is used to
    train a base \texttt{Prompt Expansion} model.
    We then align with our downstream text-to-image model to create the \texttt{PE: \rft} model  (Sec \ref{sec:rft}).
    \textit{(Inner)}
    An example of the Image-Text inversion (using COCA-Interrogator) generating an expanded prompt containing
    {\color{ci1} caption}, 
    {\color{ci2} art form}, 
    {\color{ci3} artist}, 
    {\color{ci4} medium}, 
    {\color{ci5} style}, and 
    {\color{ci6a} other flavou}{\color{ci6b}rs / objects / de}{\color{ci6c}scriptors}.
    Subsequently, the prompt is mapped by successive few-shot prompting inference calls into shorter and more abstract queries that are each paired with the prompt to construct the dataset.
    }
    \label{fig:flow}
\end{figure*}

The final step in dataset preparation is to
compute a range of potential user queries that are suitable to map to the inverted text (prompt).
We use few-shot prompting with FLAN-PaLMChilla 62B~\citep{chung2022scaling}
to generate successively shorter queries and longer prompts.
The model receives few-shot prompting sets of long prompts mapped to short queries as examples.
The few-shot prompts are in the format \texttt{\{prompt : query\}}, and examples of these pairs can be seen in Figure \ref{fig:flow} and Table \ref{tab:examples_augmentations} as well.
For each prompt from Image-to-Text inversion,
the few-shot prompt examples are prepended before the prompt as context,
and a corresponding query is generated by the text-to-text Model.

We extract a range of different queries that can be mapped to the expanded prompt,
and use few-shot prompting to generate queries that are
abstract, concrete, short-length, medium-length, or long-length.
Appendix \ref{sec:query_types} has further details on how the query types are generated (e.g. grounded queries, eliciting specificity).
This results in a Prompt Expansion Dataset of 600k \texttt{\{query:prompt\}} pairs.
We perform a 70-20-10 train-val-test split,
and split the train set 50-50 for base and re-fine-tuning.

\section{Prompt Expansion Model}

We describe the two stages to train the Prompt Expansion model:
(i) we train a base Prompt Expansion model on the Prompt Expansion Dataset;
then 
(ii) we re-fine-tune the base model with respect to the downstream text-to-image model.

\subsection{Base Model}
\label{sec:pemodel}
Our Prompt Expansion model is a text-to-text generation model trained to map query text to expanded prompt text with an architecture based on the PaLM 2 language model family~\citep{anil2023palm}. PaLM 2 is a decoder-only transformer-based architecture trained with the UL2 objective~\citep{tay2023ul}. We train a PaLM 2 1B parameter model with prompt-tuning \citep{lester-etal-2021-power}, 
after evaluating different model configurations, as described in  Table \ref{tab:model_config}. We chose this relatively small size for the base architecture, as it needs to serve as a front-end to a complex high-latency text-to-image model such as Imagen \citep{imagen-paper}, and thus needs to be low-resource/latency to make the entire pipeline usable. For the base dataset, we use a 50\% split of the Prompt Expansion Dataset described in Section \ref{sec:pemdataset}, consisting of 300k \texttt{\{query:prompt\}} examples.  

\subsection{Re-fine-tuning}
\label{sec:rft}

After training the base model, we observe that it may generate prompts that the text-to-image model cannot generate good images for. 
The main reason for this is that the expanded prompts generated by the base Prompt Expansion model are based on the alignment between text and images favored by the COCA image-to-text inversion model.
Therefore, we propose a general \textit{re-fine-tuning} procedure:
given a target behavior for the model, 
re-fine-tuning
filters for
expanded prompts generated from the base model that align with the target behaviour.
To align with the text-to-image model, 
we construct a dataset where the expanded prompts are aligned closer to the behavior of the downstream text-to-image model.

For queries in the remaining 50\% split of the Prompt Expansion dataset,  we generate expanded prompts from our base model, which are then input to the downstream text-to-image model (Imagen~\citep{imagen-paper} in our experiments). We score these images using a weighted average of the query-image embedding distance and prompt-image embedding distance (See Appendix \ref{sec:probe} for details) and filter out 
\texttt{\{query:prompt\}} pairs whose scores are below a fixed threshold.
We then continue re-fine-tuning from the base model checkpoint using only these filtered \texttt{\{query:prompt\}} pairs, 
thus producing a \texttt{PE: \rft} model which is optimized to return expansions of the query and flavors that the text-to-image model can faithfully generate high quality images for.

\begin{figure*}[t]
    \centering
    \hspace*{-2.9cm}
    \includegraphics[width=1.45\textwidth]{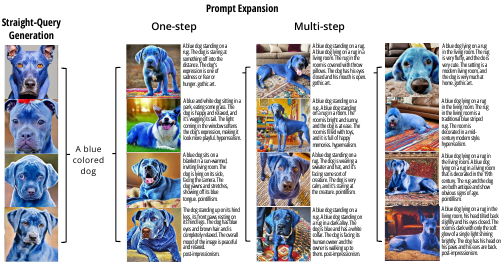}
    \caption{
    Tree of Multi-Step Prompt Expansion:
    A comparison of images generated of straight-query generation \textit{(left)} and Prompt Expansion \textit{(right)}. 
    The first image from each step is further expanded upon,
    in-line with the implementation described in Appendix \ref{sec:multistep}.
    We observe diversity in pose, action, point-of-view, background, colours. 
}
    \label{fig:}
\end{figure*}

\section{Controllable Generation}

\subsection
{Prefixes for Controlled Prompt Expansion}
Till now, we have presented our approach for building a model for generic use cases of Prompt Expansion. However, it is often the case that the user or application designer would like to control the direction of the Prompt Expansion strategy towards e.g. adding more flavors or adding specific kinds of diverse details. To support these use cases, we implement a controllable version of our Prompt Expansion model that can be directed to produce specific kinds of expansions by prepending the query with one of 8 supported \emph{Prefixes}. For example, we can direct the model to produce flavors only using the \texttt{FLV} prefix, or to iteratively expand the original query for interactive multi-step prompt expansion scenarios with the \texttt{MSTP} prefix. A few examples of controlled generation are shown in Table \ref{tab:examples_augmentations} and the full list of supported flavors are in Table \ref{tab:prefixes}. To train the \texttt{PE: Multi-Prefix} model, we begin with the Prompt Expansion dataset of Section \ref{sec:pemdataset}.
Each \texttt{\{query:prompt\}} pair is assigned with an appropriate prefix.
During few-shot prompting, some queries / prompts were formatted as abstract or detailed, thus these prefixes are known (e.g. \texttt{ABST}, \texttt{DTL}).
Some prefixes (e.g. \texttt{RFT}, \texttt{MSTP}) are also known as their \texttt{\{query:prompt\}} pairs are synthesized.
Some \texttt{\{query:prompt\}} pairs need to be classified to its prefix, for example \texttt{HAST} prefixes are assigned to prompts whose images return good aesthetics.
Prefix assignment resulted in a new version of the Prompt Expansion dataset with every query prepended with a prefix; and this was used to fine-tune and train the \texttt{PE: Multi-Prefix} model.

\begin{table*}[t]
\centering
\caption{
Aggregate Prompt Expansion performance.
}
\label{tab:aggregate}
    \begin{subtable}[ht]{\columnwidth}
        \centering
        \tiny
        \resizebox{\textwidth}{!}{%
        \begin{tabular}{l|ccc}
        \hline \hline
        \textbf{Method}
          & \textbf{Aesthetics (MUSIQ-AVA) $\uparrow$} & \textbf{Text-Image Alignment (COCA(q, I(p))) $\uparrow$} & \textbf{Diversity ($\sigma_{p}$)) $\uparrow$} \\ \hline
Straight-Query Generation 	& 	5.121 $\pm$ 0.519 &	0.125 $\pm$ 0.0147 &	0.00582 $\pm$ 0.00275 \\  \hline
Few-shot Prompting & 5.295 $\pm$ 0.549 &	0.114 $\pm$ 0.0199 &	 0.00726 $\pm$ 0.00339  \\  \hline
Prompt Expansion & 5.225 $\pm$ 0.585 &	0.120 $\pm$ 0.0175 &	0.00720 $\pm$ 0.00341 \\  \hline
\rowcolor{blue!10} PE: \rft & 6.185 $\pm$ 0.474 &	0.113 $\pm$ 0.0199 &	0.00746 $\pm$ 0.00354 \\  \hline\hline
PE: Multi-Prefix &  5.712 $\pm$ 0.616 &	0.125 $\pm$ 0.0157 &	0.00624 $\pm$ 0.00297  \\  \hline
PE: Prefix Dropout &  5.410 $\pm$ 0.622  &	0.121 $\pm$ 0.0156 &	0.00634 $\pm$ 0.00304  \\  \hline
\hline
        \end{tabular}%
        }
    \end{subtable}

\end{table*}

\subsection
{Prefix Dropout for Generic Prompt Expansion}
With the Multi-Prefix dataset in hand, we explored the possibility of using controllable generation hints to improve the performance on the generic Prompt Expansion task. The idea is to initialize the training of the model using controlled generation and then gradually shift its behavior over the course of the training to guess an appropriate prefix for a given query and generate the matching expansions e.g. for highly abstract queries such as \emph{"Undying Love"}, the model's behavior should match that of the \texttt{ABST} prefix (See Table \ref{tab:examples_augmentations}).  This is achieved through a novel curriculum learning technique we call \texttt{Prefix Dropout} 
where we start with the prefix-annotated dataset described above, but over the course of training steadily increase the percentage of examples where the prefix is randomly removed or dropped-out from the query
starting from a 0.4 dropout rate to 1.0. This yields the  \texttt{PE: Prefix Dropout} model which can be compared to our base and re-fine-tuned models as a candidate for generic Prompt Expansion.

\subsection
{Multi-Step Prompt Expansion}
Exploration can be a multi-step process.
After the user's query returns a set of expanded prompts, 
the user can select amongst the prompts, 
and this prompt is 
fed back into the Prompt Expansion model.
This allows users to iterate on the expanded prompts without the need for manually prompt engineering the text.
Using \texttt{PE: \rft}, we generate expanded prompts on held-out queries, and iteratively generate prompts upon the previous step's prompts.
This results in multi-step training data of expanded prompts to next-step expanded prompts.
We re-fine-tune the Prompt Expansion model with the multi-step data prepended with the \texttt{MSTP} prefix.

\section{Experiments}
\label{sec:metrics}

We conducted evaluations of the different flavors of PE described above, focusing mainly on their ability to generate diverse and aesthetically pleasing images, without signifcant semantic drift in the prompt. We conducted these using both automatic metrics and a human evaluation with paid interns. In the rest of this section, we describe the experimental setup, task design, and metrics.

\subsection{Evaluation Setup}

\textbf{Evaluation set.}
For both automatic and human evaluation experiments, we sample queries from 2 sources: (1) n=200 prompts from the PartiPrompts (PP) \citep{yu-parti} dataset of prompts that is designed to represent different domains and features of language (such as counting, negation etc); (2) n=500 prompts from a novel test set of potential queries constructed by applying the PE dataset generation process of Section \ref{sec:pemdataset} (Text-Inversion + Query/Prompt Extraction) to \waa{} (WA) \citep{chen2023pali, jia2021scaling} images. Queries are categorized as
abstract,
concrete,
short (<4 words),
medium length (4-7 words),
or long length (>7 words).
For WA, we obtain prompts of different lengths by selecting prompts from different steps of the prompt shortening process (Sec \ref{sec:pemdataset}).

\noindent
\textbf{Models.}
We evaluate three variations of Prompt Expansion: (1) A \texttt{Few-shot Prompting} baseline, where we few-shot prompt a FLAN-PaLMChilla 62B \citep{palm-2022} model to generate expanded prompts given examples of query-prompt mappings.
This is similar to the setup in Section \ref{sec:pemdataset},
except we swap the input and output to generate prompts from queries.
(2) The base \texttt{Prompt Expansion} model which is constructed following all the steps in Section \ref{sec:pemodel}, except for re-fine-tuning. (3) A model trained using the full \texttt{Prompt Expansion} pipeline, which we call \texttt{PE: \rft}.
We compare the results of these  to \texttt{Straight-Query Generation}, where the text-to-image generation model is given the original query as input.
All our prompts are given as input to Imagen \citep{sahariac-imagen} and the resulting images are evaluated using the metrics below.

\subsection{Metrics for Automatic Evaluation}

We use 3 metrics to evaluate the images generated from the prompts returned by the models described above:
\begin{itemize}
\item 
For \textbf{aesthetics}, we use MUSIQ (pre-trained on the AVA dataset) \citep{ke2021musiq}.
A higher score is better, and is primarily used for evaluating distortions in photorealistic images (cartoon images, pixel art, and paintings tend to receive lower scores). 
\item 
For \textbf{diversity}, we use the variance($\sigma_p$) of the COCA \citep{yu2022coca} image embeddings of the generated images.
\item 
For \textbf{text-image alignment}, we use the cosine similarity between the COCA text embeddings of the query text against the COCA image embeddings of the generated image, similar to CLIP-score \citep{hessel2021clipscore}.
For a query $q$ used to generate an expanded prompt $p$, which is then used to generate its corresponding image $I(p)$,
to measure the COCA embedding distance $\texttt{COCA}(\cdot, \cdot)$
we denote the 
COCA score 
between query text and expanded prompt image as 
$\texttt{COCA}(q, I(p))$,
and between prompt text and expanded prompt image as $\texttt{COCA}(p, I(p))$.
\end{itemize}

\subsection{Task Design for Human Evaluation}

We perform a side-by-side (SxS) evaluation task, where we show raters a pair of images side by side for a given text query, where each image is generated by a prompt from one PE model or the other, and ask the rater to pick the best one. We do this for 2 separate criteria, aesthetic preference and text-image alignment (details of the task design in Appendix \ref{sec:rater}). We split our query set into two subsets of 350 queries each to be used in the following settings:  
\begin{enumerate}[leftmargin=*]
\item
\textbf{Random vs. Random (1x1)}: A random image generated from each prompt is compared SxS. 
\item
\textbf{Best vs. Best (4x4)}: For each prompt, we generate 4 images using the text-to-image model, ask raters to select the best one for each prompt in separate tasks, and then ask a fresh rater to compare the selected images SxS.
\end{enumerate}

If two images contain all the attributes and details specified by the query, there are no further distinctions from which a rater may rate one image as being more aligned to the query than the other. Thus, for text-image alignment we allow the rater to judge the SxS rating task as "Equivalent" when they find no significant difference between the images, but for Aesthetics they are required to select one of the images as better. We analyze the results using Prompt-win rates vs Query-win rates (the percentage of time each rater chose the prompt-generated image over the query-generated image), along with reporting Equivalent-rate for text-image alignment. 

\section{Results and Discussion}
\label{sec:results}

\begin{figure*}[t]
    \centering
    \hspace*{-2.5cm}
    \includegraphics[width=.25\textwidth]{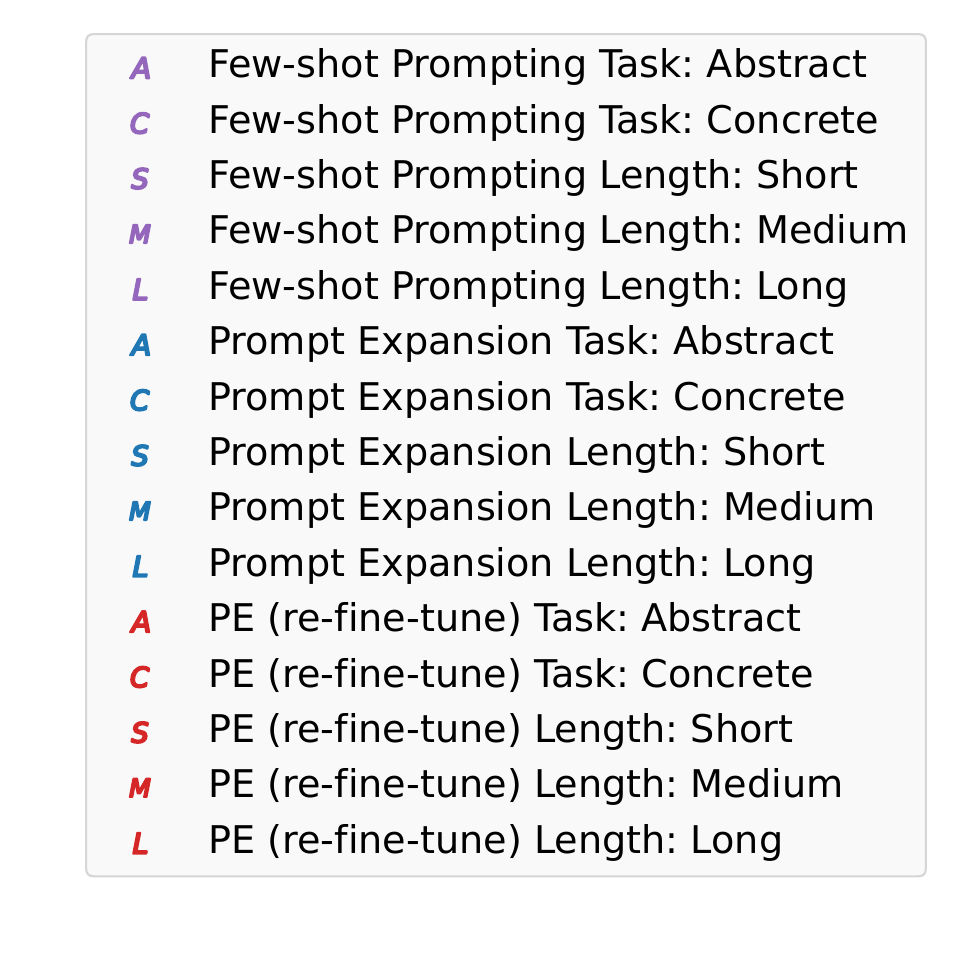}
    \includegraphics[width=.3\textwidth]{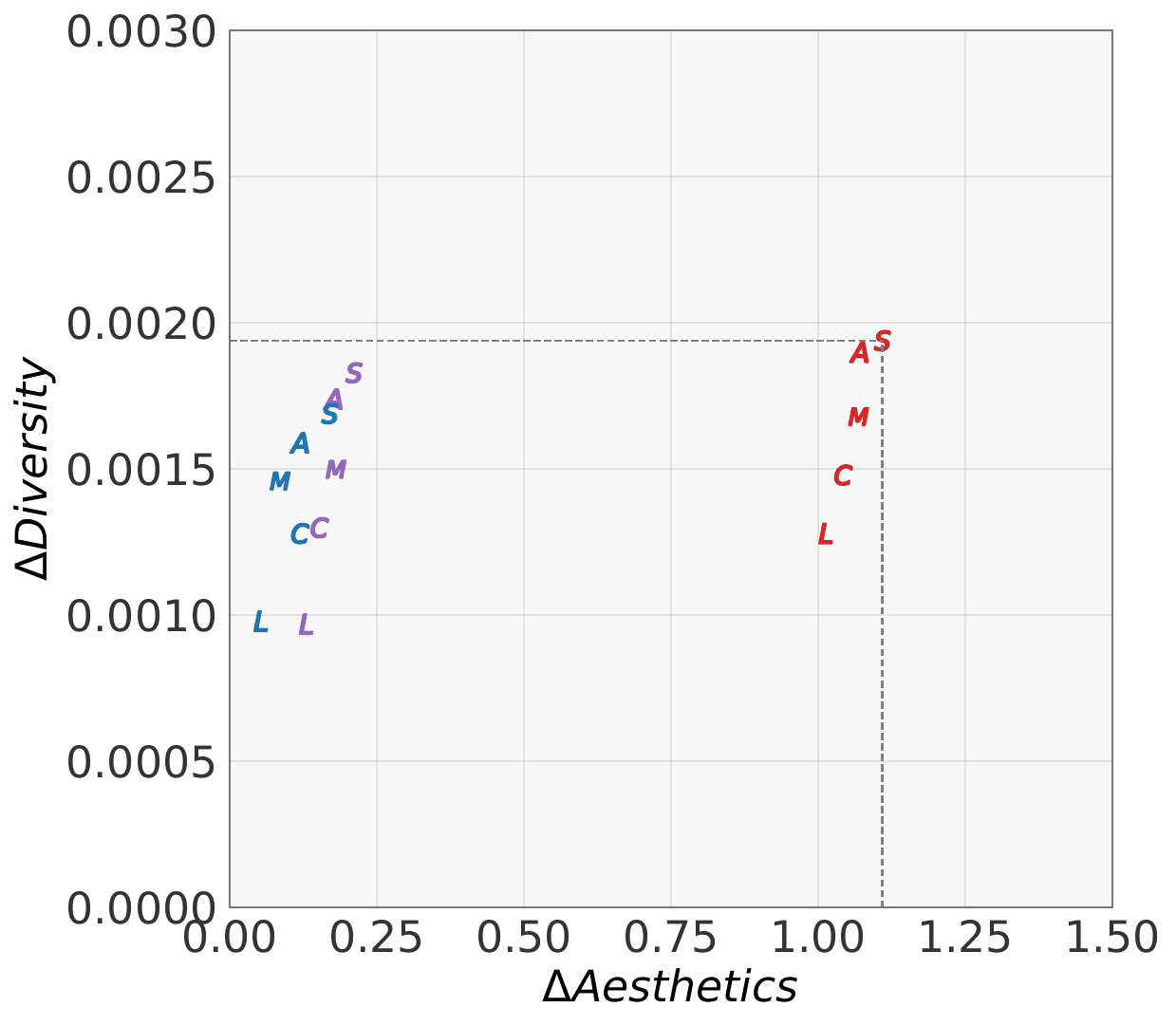}
    \includegraphics[width=.3\textwidth]{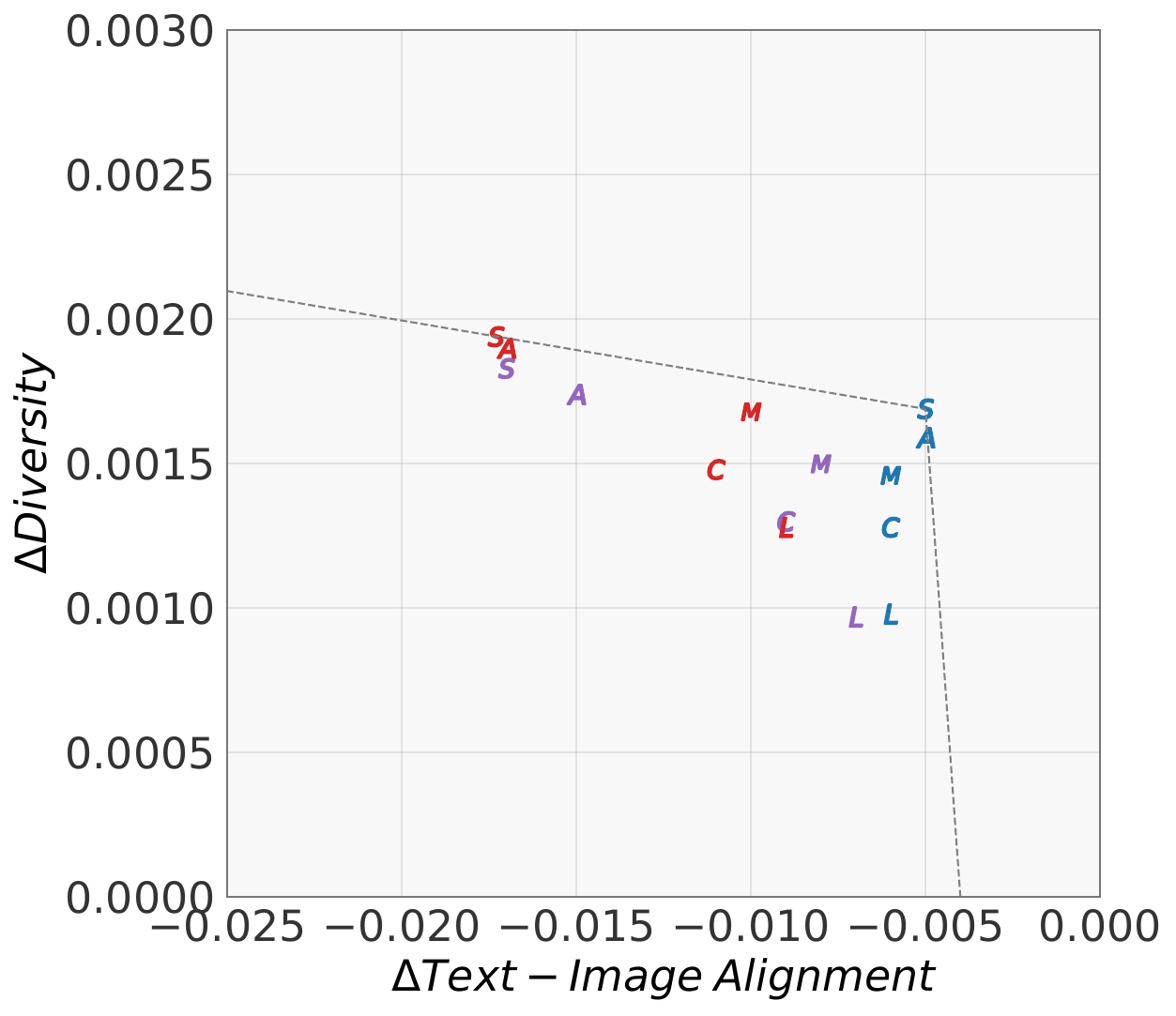}
    \includegraphics[width=.28\textwidth]{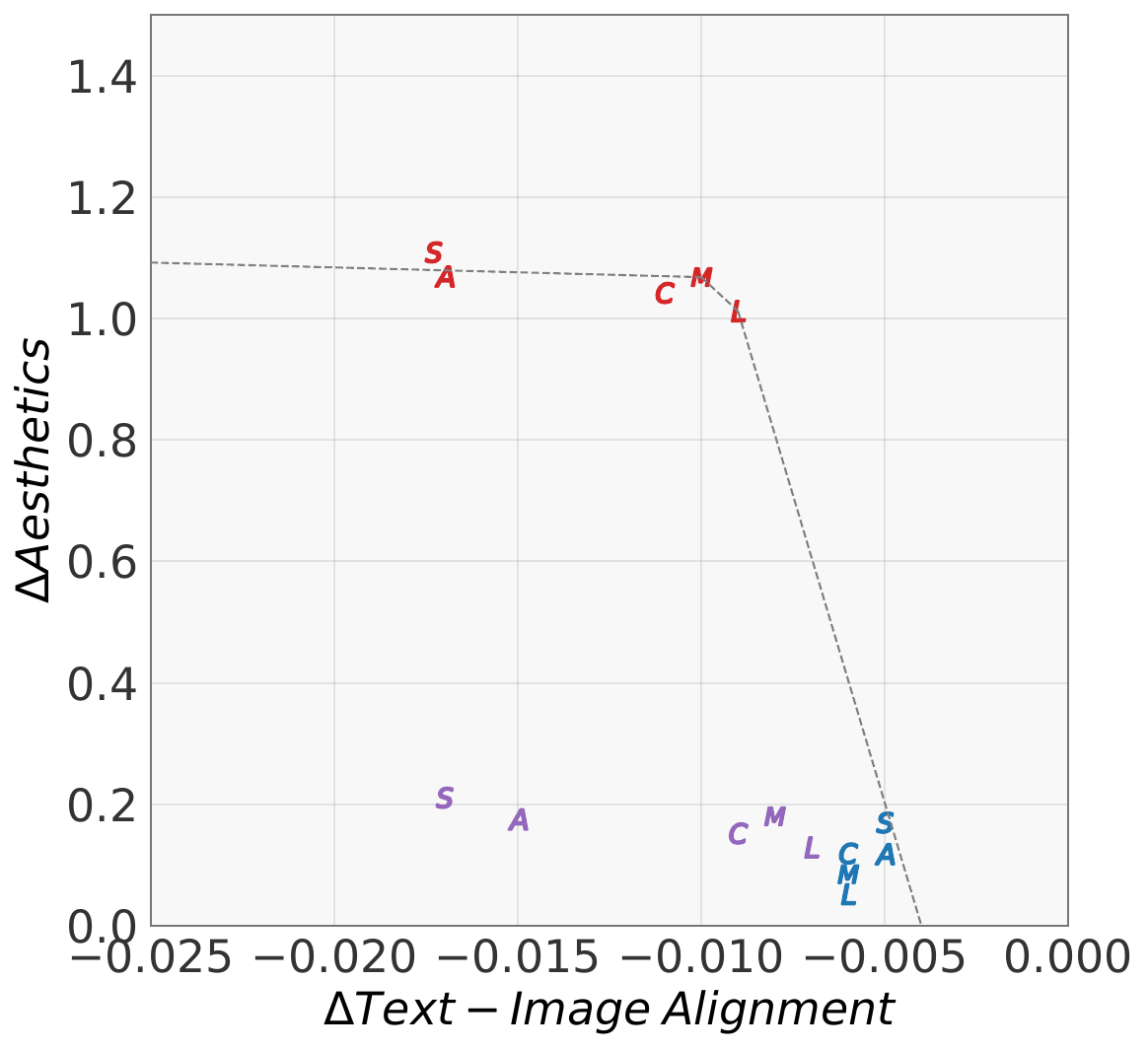}
    \\
    \hspace*{-2.5cm}
    \includegraphics[width=.25\textwidth]{assets/legend.pdf}
    \includegraphics[width=.3\textwidth]{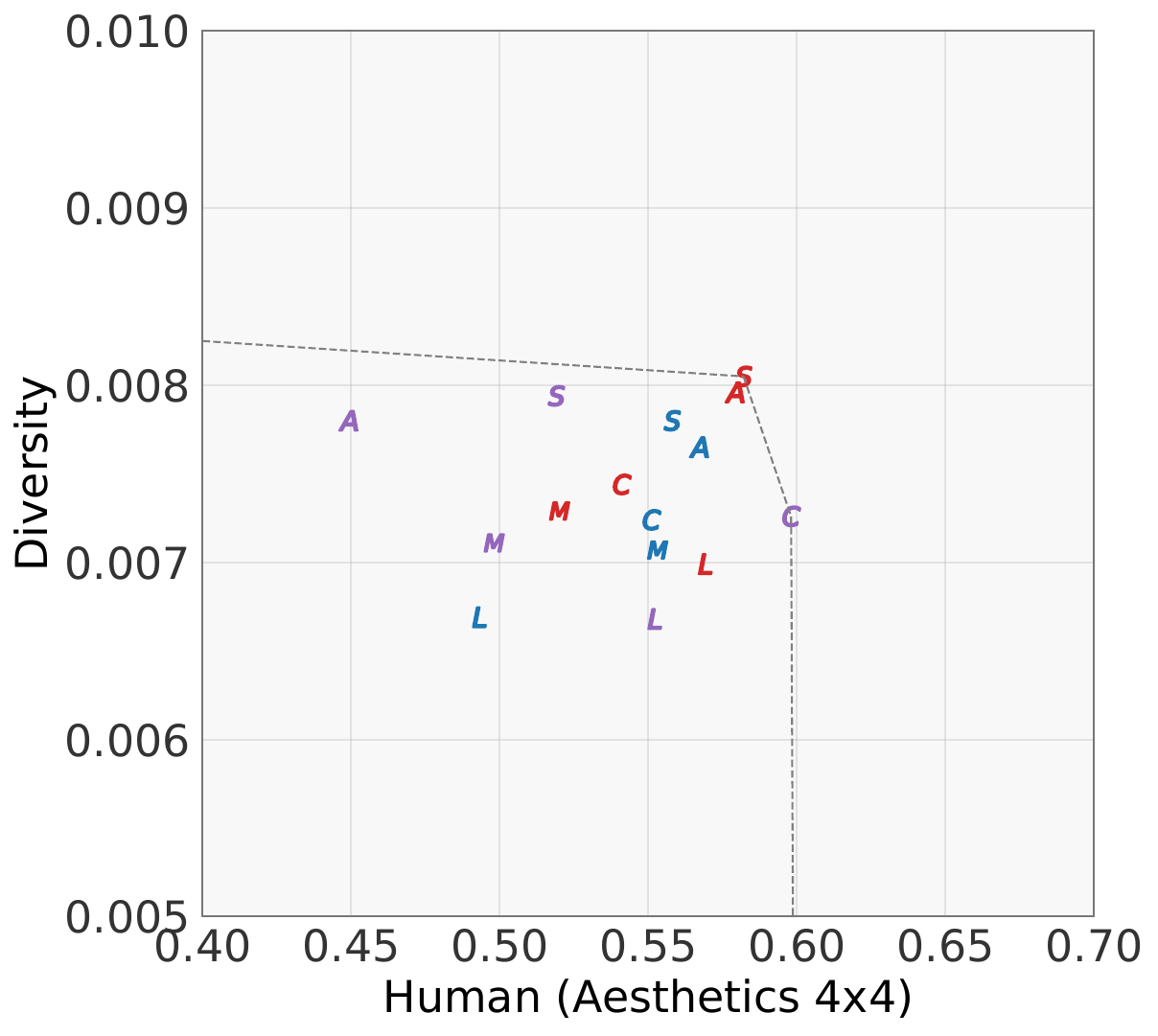}
    \includegraphics[width=.3\textwidth]{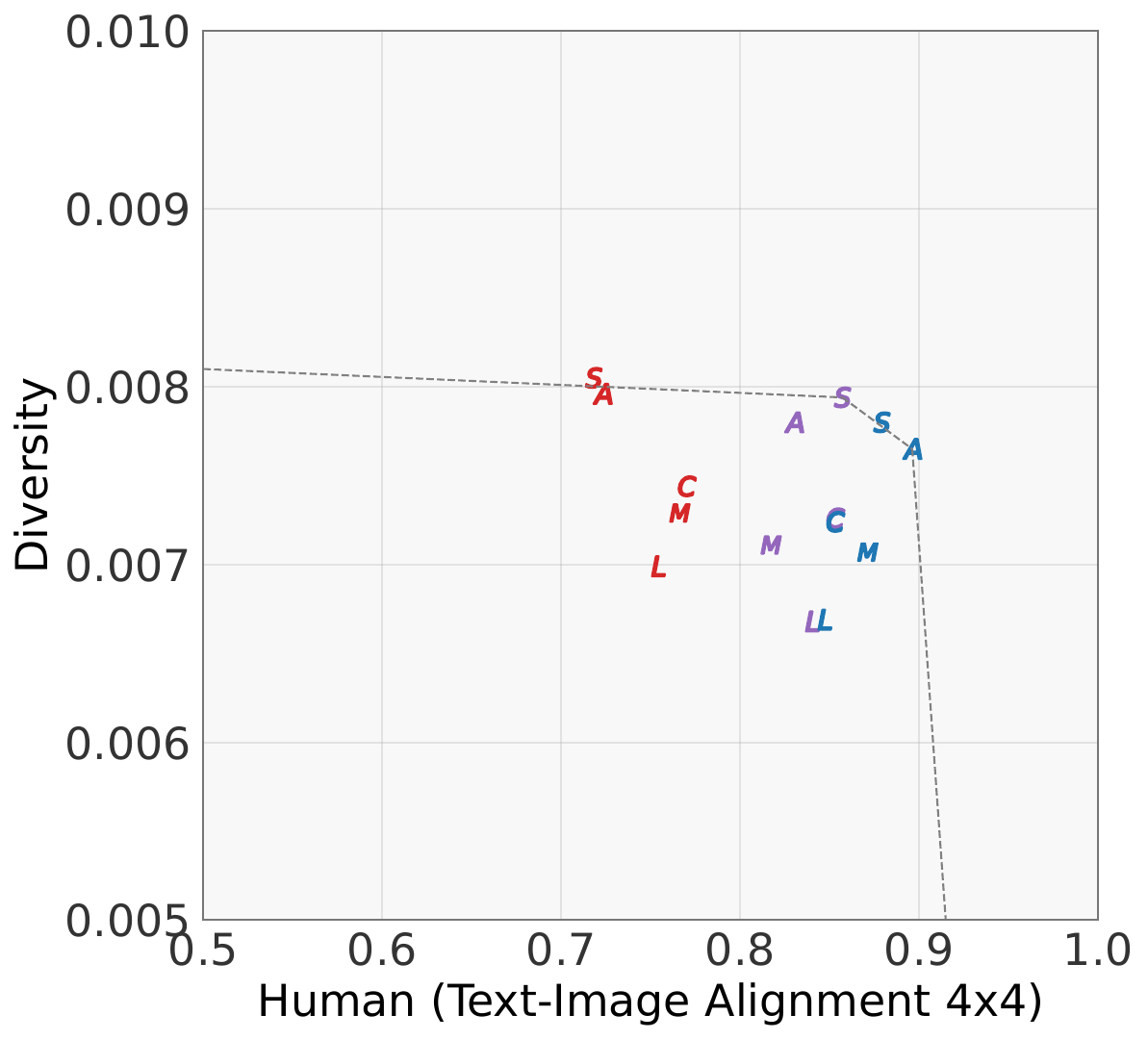}
    \includegraphics[width=.29\textwidth]{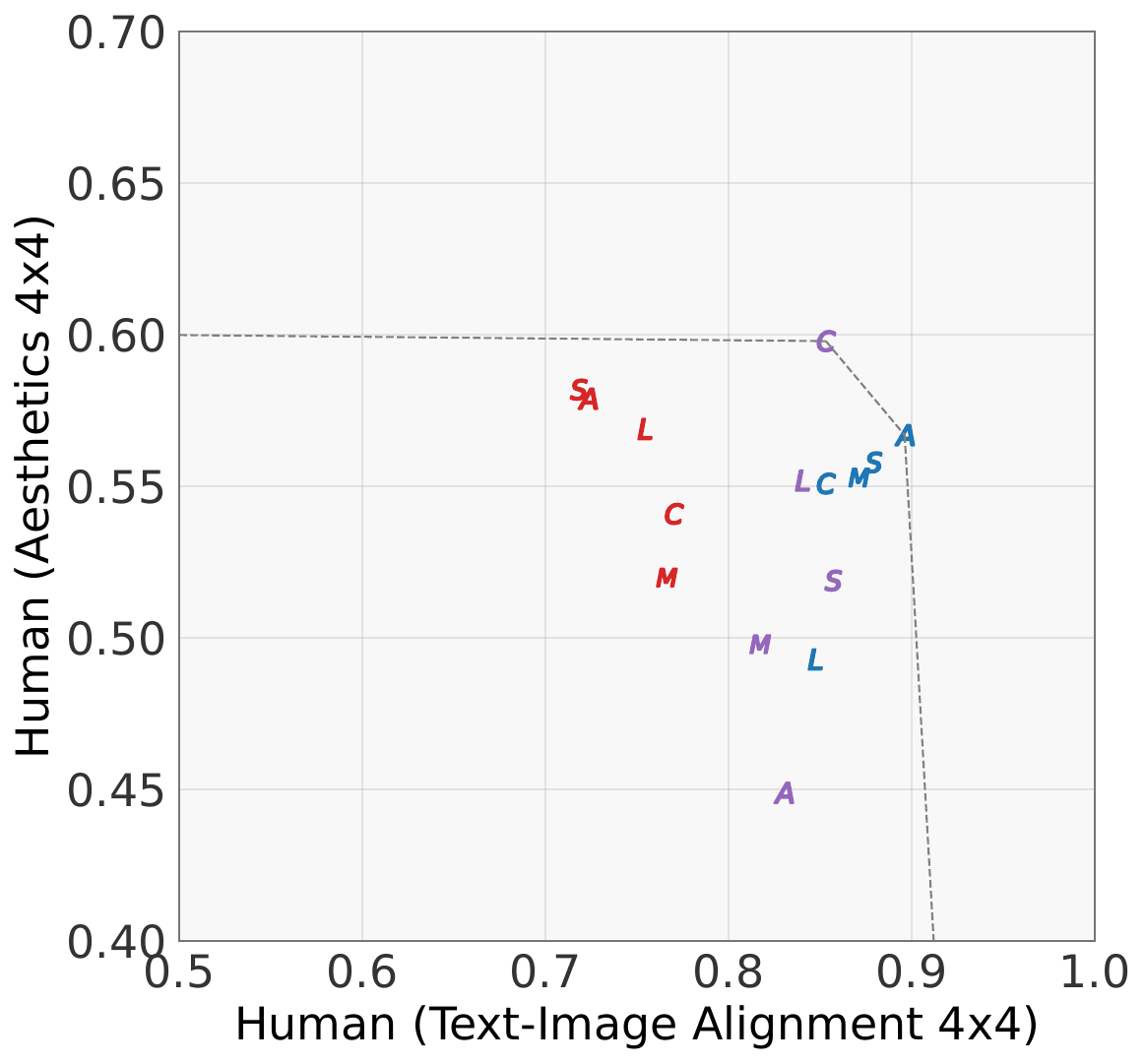}
    \caption{
    We compare (\texttt{\{Few-shot Prompting, Prompt Expansion, and PE (\rft)\} - Straight-Query Gen.}).
    For all automatic evaluation metrics, we plot the difference $\Delta \textnormal{\textit{metric}} = \textnormal{\textit{metric}}_{\textnormal{\textit{PE}}} - \textnormal{\textit{metric}}_{\textnormal{\textit{SQG}}}$.
    For all human rater evaluation metrics, the rater study already compares relative to Straight-Query Generation. 
    For Human (Consistency) values, we plot Prompt+Equivalent scores.
    For completeness, we provide all numbers in Figures \ref{tab:variations}
    and \ref{fig:hres2}-\ref{fig:hres1}.
    }
    \label{fig:tradeoff}
\end{figure*}

From the results of the automatic (Table \ref{tab:aggregate}) and human evaluation (Figures \ref{fig:tradeoff}, \ref{fig:hres2}, and \ref{fig:hres1}), we derive some insights relevant to the Research Questions raised in Section \ref{sec:intro}.

\noindent
\textemdash
\textbf{Does Prompt Expansion increase diversity of the generated images?}
Table \ref{tab:aggregate} shows a small but consistent increase in diversity scores between our various Prompt Expansion models and the Straight-Query baseline.  
Though small compared to the confidence interval, the gains are similar across all systems and hold consistently across query types (see Table \ref{tab:variations}), increasing our confidence in the positive impact of PE on diversity. 
We observe \texttt{PE: \rft} has marginally higher diversity than other PE models, both in aggregate (Table \ref{tab:aggregate}) and  per query type (Table \ref{tab:variations}).

We also found it possible to further increase overall diversity by using techniques such as temperature sampling and post-hoc filtering (See Table \ref{tab:div_ablations} for results).
Qualitatively, this increase in diversity manifests in 
topical, semantic, and expressive diversity, and in
fairness considerations such as age, gender, and culture (See the examples in Table \ref{tab:fairness}). 

\noindent
\textemdash
\textbf{Does prompt optimization lead to better aesthetics?}
The \texttt{PE: \rft} model shows a significant increase in the aesthetic score vs. Straight-Query generation, while the gain for the other PE models are more modest. We conclude therefore, that the text-to-image model-relative parts of our pipeline are critical to improving aesthetic values. 
This conclusion is further strengthened by examining the rater preference results (Figure \ref{fig:hres2}), where we see that re-fine-tuning consistently achieves higher win-rates over straight-query baselines (scoring $>0.52$ across all query types) than the other models. For the 1x1 setting, this can be entirely attributed to the prompt being optimized during re-fine-tuning for higher aesthetics. A significant source of the gains in aesthetics is due to the Prompt Expansion model learning which flavors the downstream text-to-image model is most responsive to (Examples of these are shown in Figure \ref{tab:flavor_probe_full}).   

\noindent
\textemdash
\textbf{Does increased diversity of samples lead to better aesthetics?}
We hypothesize that in addition to the benefits of tuning the expanded prompt to generate higher quality images, there is an additional gain of quality achieved when images from a single prompt are more diverse. With higher diversity, the likelihood of a highly aesthetically pleasing image being generated and selected by the user goes up. To demonstrate this effect, we compare the difference in Prompt-win rates between the 4x4 (where the best images from each system are compared) and 1x1 settings, see Figure \ref{fig:hres2}. The Prompt-win rates are consistently larger across all systems in the former, showing the aesthetic benefits of increased diversity at both the prompt and image level.  

\begin{table*}[t]
\centering
\begin{minipage}{\textwidth}
\centering
\includegraphics[width=\textwidth]{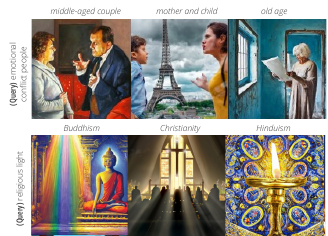}
\vspace{0.5cm}
\end{minipage}
\begin{minipage}{\textwidth}
\centering
\resizebox{\textwidth}{!}{%
\begin{tabular}{p{2cm}p{3cm}p{10.5cm}}
\hline \hline
\textbf{Aspect}
& \textbf{Query} 
& \textbf{Prompt} \\ \hline
Religion
& religious light & 
\textemdash {\color{purple}An iconic statue of Buddha in a buddhist monastery.} The statue is majestic and dominating in its surroundings. The statue represents peace and harmony, and the monastery is filled with monks and monks. art deco.
\\
&& 
\textemdash {\color{purple}A depiction of a church with many people worshipping in it during evening service.} The church is full of sunlight and the people are all dressed up in black. The church is ornate and the people are singing and dancing.
\\\hline 
Culture
& 
A woman wearing a long, flowing orange and white dress. 
& \textemdash A beautiful portrait of a woman in a flowing orange and white dress. {\color{purple}The dress is long and flowing, with a deep v-neck and flowing sleeves.} The background is a bright sunrise, with a lone cloud in the sky. The overall mood of the work is nostalgic and romantic. futurism.
\\
&& 
\textemdash {\color{purple}A portrait of a young woman wearing a traditional Indian wedding dress.} The woman is smiling and looking over at her husband. The dress is flowing, flowing, flowing, and the lighting is soft and romantic. futurism.
\\\hline 
Age \& Gender
& angular eyebrows & 
\textemdash {\color{purple}A black-and-white photograph of a man's eyes with thick, angular eyebrows.} The photograph is at least two decades old, and the man is most likely still alive. hyperrealism.
\\
&& 
\textemdash {\color{purple}A young woman with wide-open eyes and an angular, sagginess to her eyebrows.} The woman is looking into the mirror, her face is radiant with confidence. trending on cg society.
\\
&& 
\textemdash {\color{purple}A young boy has angular eyebrows with dark hair.} The boy is smiling, and he is wearing a dark shirt and dark pants. The boy is standing in the grass, and he is looking at the sky with a dark grin on his face. post-impressionism.
\\\hline 
\hline
        \end{tabular}%
}
\end{minipage}
\caption{
\textbf{Fairness from diversity}:
Prompt Expansion generates prompts that introduce variations across religion, culture, and demographics at the textual level. We highlight in these examples, the sentences that are indicative of the variation. 
    Despite no explicit training for specific topical diversity,
    Prompt Expansion responds to a query on religion by showing imagery from different religious backgrounds from Buddhism, Christianity, and Hinduism.
    It responds to another underspecified query "emotional conflict people" by showing people in different situations of emotional conflict and of varying age demographics.
    For another example,
    Prompt Expansion still finds small ways to introduce cultural variations that remain faithful to a long and detailed query, such as describing dresses from different global contexts. 
}
\label{tab:fairness}
\end{table*}

\clearpage

\noindent
\textemdash
\textbf{Does Prompt Expansion balance between aesthetics and text-image alignment? }
An important observation is that while an expanded prompt can be designed to preserve all the semantics of the original query (by ensuring no details are lost etc.), it can never be expected to increase the degree of alignment of the image to the original intent of the user as expressed by the query. Thus, the main goal of Prompt Expansion with respect to text-image alignment is to minimize the decrease of alignment scores while optimizing for aesthetics and diversity. Thus, we expect to see a tradeoff between alignment against and aesthetic/diversity scores.

Indeed, in Table \ref{tab:aggregate}, we see that there is a minimal drop in the alignment scores of our PE models, inversely correlated with the other 2 metrics of diversity and aesthetics. The alignment score drop however, is well within the confidence intervals, showing that the semantics of the expanded prompt do not stray too far from the meaning of the original query. As further evidence of this, we turn to the human rater evaluation for text-image alignment (Figure \ref{fig:hres1}) where see large Equivalent-rates in the comparisons, and comparable win-rates for prompt and query. For example, in the 1x1 setting for the re-fine-tuning model, we see a 70\% equivalent-rate and prompt/query win-rates of 15\% each.    

\begin{table*}[t]
\centering
	\begin{minipage}{\columnwidth}
		\centering
     \centering
     \captionof{figure}{
Human Rater results for Aesthetics in 1x1 and 4x4 settings.
}
    \begin{subfigure}[t]{.3\columnwidth}
         \centering
        \includegraphics[width=\columnwidth]{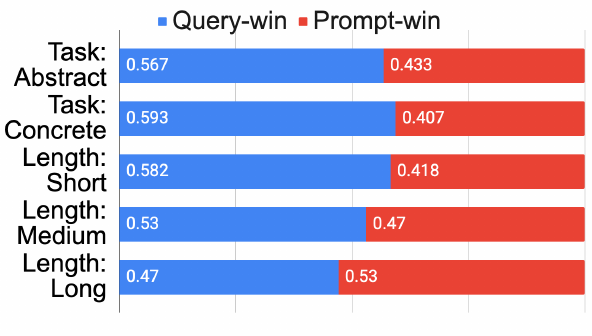}
         \caption{Human Evaluation (1x1 Aesthetics, Few-Shot Prompting)}
     \end{subfigure}
     \begin{subfigure}[t]{.3\columnwidth}
         \centering
        \includegraphics[width=\columnwidth]{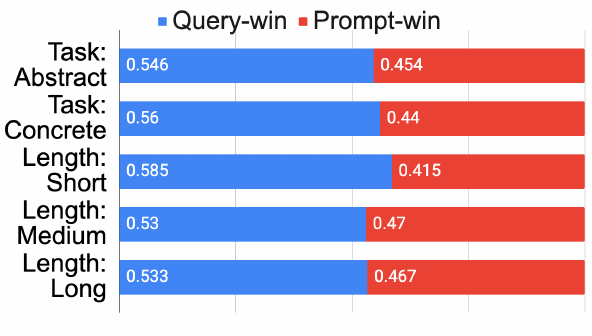}
         \caption{Human Evaluation (1x1 Aesthetics, Prompt Expansion)}
     \end{subfigure}
     \begin{subfigure}[t]{.3\columnwidth}
         \centering
        \includegraphics[width=\columnwidth]{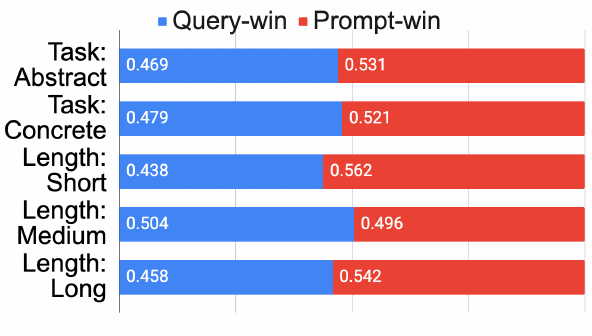}
         \caption{Human Evaluation (1x1 Aesthetics, PE: re-fine-tuning)}
     \end{subfigure}
    \begin{subfigure}[t]{.3\columnwidth}
         \centering
        \includegraphics[width=\columnwidth]{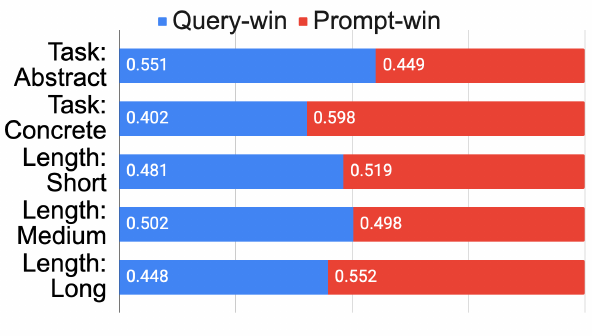}
         \caption{Human Evaluation (4x4 Aesthetics, Few-Shot Prompting)}
     \end{subfigure}
     \begin{subfigure}[t]{.3\columnwidth}
         \centering
        \includegraphics[width=\columnwidth]{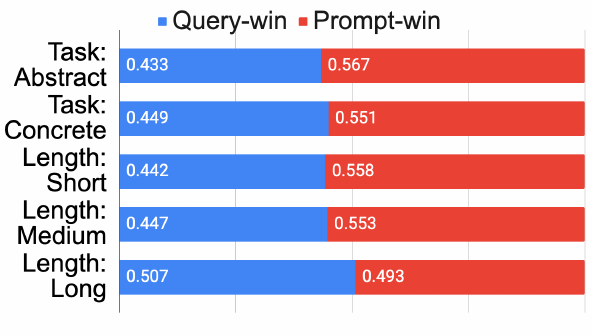}
         \caption{Human Evaluation (4x4 Aesthetics, Prompt Expansion)}
     \end{subfigure}
     \begin{subfigure}[t]{.3\columnwidth}
         \centering
        \includegraphics[width=\columnwidth]{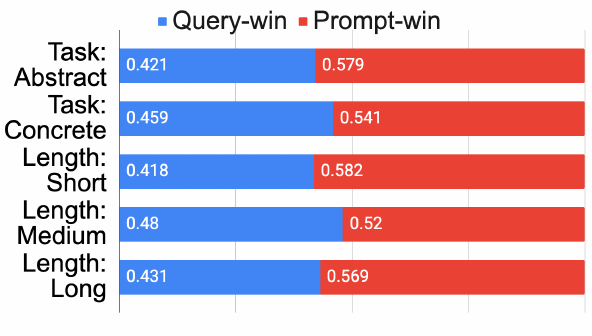}
         \caption{Human Evaluation (4x4 Aesthetics, PE: re-fine-tuning)}
     \end{subfigure}
\label{fig:hres2}
	\end{minipage}
	\begin{minipage}{\columnwidth}
		\centering
     \centering
     \captionof{figure}{
Human Rater results for Text-Image Alignment in 1x1 and 4x4 settings.
}
     \begin{subfigure}[t]{.3\columnwidth}
         \centering
         \includegraphics[width=\columnwidth]{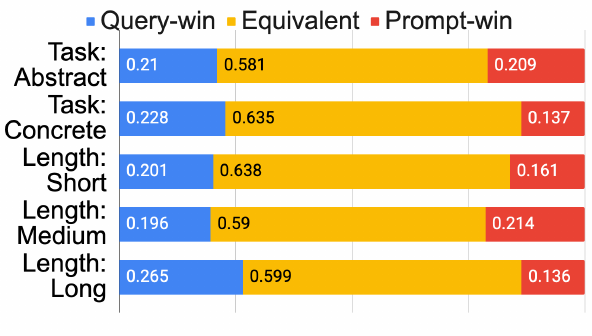}
         \caption{Human Evaluation (1x1 Text-Image Alignment, Few-Shot Prompting)}
     \end{subfigure}
     \begin{subfigure}[t]{.3\columnwidth}
         \centering
         \includegraphics[width=\columnwidth]{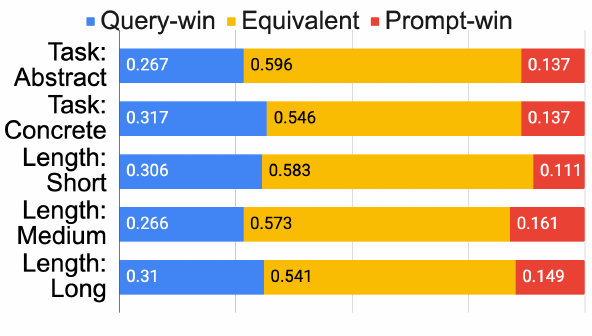}
         \caption{Human Evaluation (1x1 Text-Image Alignment, Prompt Expansion)}
     \end{subfigure}
     \begin{subfigure}[t]{.3\columnwidth}
         \centering
         \includegraphics[width=\columnwidth]{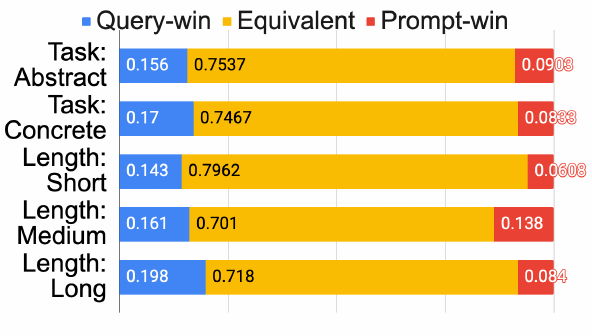}
         \caption{Human Evaluation (1x1 Text-Image Alignment, PE: re-fine-tuning)}
     \end{subfigure}
    \begin{subfigure}[t]{.3\columnwidth}
         \centering
         \includegraphics[width=\columnwidth]{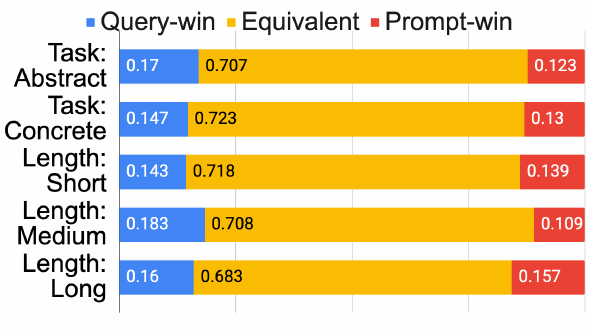}
         \caption{Human Evaluation (4x4 Text-Image Alignment, Few-Shot Prompting)}
     \end{subfigure}
     \begin{subfigure}[t]{.3\columnwidth}
         \centering
         \includegraphics[width=\columnwidth]{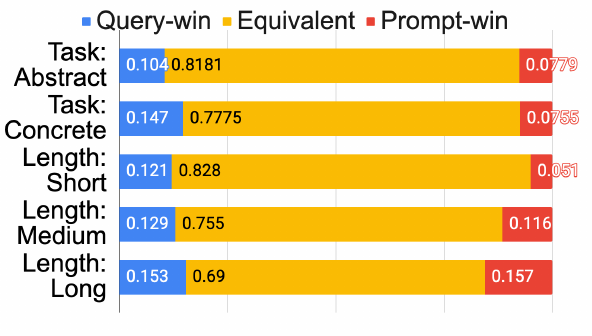}
         \caption{Human Evaluation (4x4 Text-Image Alignment, Prompt Expansion)}
     \end{subfigure}
     \begin{subfigure}[t]{.3\columnwidth}
         \centering
         \includegraphics[width=\columnwidth]{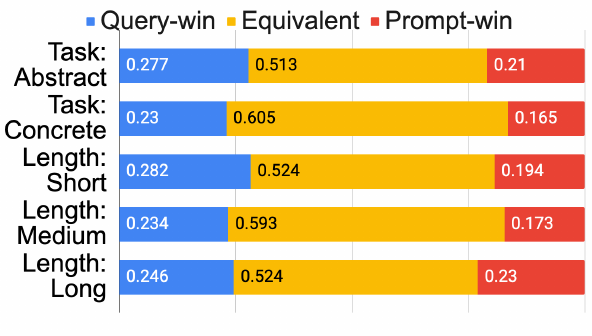}
         \caption{Human Evaluation (4x4 Text-Image Alignment, PE: re-fine-tuning)}
     \end{subfigure}
\label{fig:hres1}
    \end{minipage}
\end{table*}

We further explore the trade-offs between diversity, aesthetics, and text-image alignment by presenting Pareto-optimality curves between the various automatic and human metrics for all the different query types in Figure \ref{fig:tradeoff}. While \texttt{PE: \rft} dominates \texttt{Prompt Expansion} in terms of Diversity-Aesthetics, we notice that the converse is true for Diversity-Alignment.
If text-image alignment is more important than diversity and aesthetics collectively, 
then it may be beneficial to omit the re-fine-tuning step.

\noindent
\textemdash
\textbf{Can controllable generation improve performance?}
In Table \ref{tab:aggregate}, 
we see that \texttt{PE: Multi-Prefix} obtains better aesthetics than our base PE model, and remarkably without any loss in text-image alignment. Note however that these results are obtained with experts annotating the test set with the optimal prefix for each query before it is input to the model. Thus, these results should be seen more as an upper-bound on how much controlled generation can benefit the generic Prompt Expansion task.  However, we also see that the \texttt{PE: Prefix Dropout} model retains much of the advantage of \texttt{PE: Multi-Prefix}, without matching the aesthetic performance of the \texttt{PE: \rft} model. It has the advantage however, of not being optimized for a particular downstream text-to-image model and is thus more generally more applicable.

\section{Conclusion}
Text-to-image generation models are capable of producing high-quality images from text prompts, but they require specialized skills and knowledge to use effectively. 
Prompt Expansion reduces the need for users to iteratively prompt-engineer and over-specify their text prompt.
Human raters find that images generated through Prompt Expansion are more aesthetically-pleasing and diverse than baselines.
Empirical metrics for aesthetic and diversity also show that Prompt Expansion outperforms comparable baselines.
In addition, the paper discusses the use of Prompt Expansion as a building block for other use cases such as multi-step adaptation and controllable generation.

\section*{Acknowledgements}

We would like to thank Jason Baldridge, Austin Waters, Roopal Garg, and Su Wang for discussions, guidance, and regular support throughout the project. 
We thank Stefan Zinke, Amnah Ahmed, Jason Wang, Sunny Mak, and Ravi Rajakumar for building the text-to-image platforms from which the \crowdsource{} dataset was collected, and for general infrastructure support.
We thank Jordi Pont-Tuset, Marisa Ferrara Boston, Alena Butyrna, Josh Lee, Max Lin, and internal product teams at Google for guidance in the evaluation and deployment of Prompt Expansion, and Isabelle Guyon and Tania Bedrax-Weiss for executive support. We thank Chin-yi Cheng for reviewing an early version of this draft.
We also thank the many colleagues working on 
the text-to-image generation effort, 
datasets, language and image models, and scalable deep learning infrastructure
within Google 
for helping to make the project a reality.

\clearpage
\bibliographystyle{utils/acl_natbib}
\bibliography{main}
\clearpage
\appendix
\section*{\centering Appendix}
\section{Types of queries}
\label{sec:query_types}

Queries can arise in different structures or user intents at test-time. 
We extract a range of potential user queries that would be mapped to the inverted text (prompt).
One intent for this 
is that a range of user queries may map to the same prompt text.
For example, the user query may be abstract and underspecified, and a Prompt Expansion model is needed to introduce ideas and potential content directions.
Or the user query may be very concrete and detailed, but needs refinement in fine-grained details or phrasing to improve image aesthetics.
The latter example is supportive of an iterative approach to image generation,
where prompt/image details are added incrementally. 

Focused on adapting Prompt Expansion to the level of image specification provided by the user, 
there are two general settings that we consider:
semantic expansion, and detailed expansion.
Semantic expansion is a Prompt Expansion setting where the user queries are abstract or underspecified, the user intent may be towards ideation or exploration, and the output prompts intend to add details and expand on the query conceptually.
Detailed expansion is a Prompt Expansion setting where the queries may be quite detailed and concrete, and the expanded prompt needs added details while still remaining faithful to the original query.
This consequently results in the construction of
additional sub-datasets (mixtures) to be added to the base Prompt Expansion dataset.
These augmentative mixtures include
abstract queries (to be used in evaluating
semantic expansion), 
detailed queries (such as grounded / specificity mixtures,
to be used in evaluating detailed expansion), etc. 

To further support these wider
range of query use cases (abstract, detailed, varying sentence length), we 
few-shot prompt to generate more training data accordingly.
For example, grounded \texttt{\{query:prompt\}} pairs require the sub-phrases in the input query to reside in the output prompt,
while
specificity \texttt{\{query:prompt\}} pairs require the format of the output prompt to begin with the query, and the remainder of the prompt is posited as the example of the query (Table \ref{tab:examples_augmentations}). 
Constructing these augmentations need to be mindful of grammatical errors, which may reduce human readability in the expanded prompts.
We also need to need to be mindful of the token distribution of the training dataset, so as to avoid skewing the model towards generating prompts on very specific topics, 
or repeating tokens (e.g. in Figure \ref{fig:bad_examples}b).
These mixtures are also train-test split, where the test splits can be used to evaluate models across varying query types.
Examples of these query types can be seen in Table \ref{tab:examples_augmentations}.

\section{Adapting to query specification}

\subsection{Controllable Generation with \texttt{Multi-Prefix}}
Not all queries are the same, and
different queries may have different requirements on what their expanded prompts should look like.
We assign different query types with a corresponding prefix (Table \ref{tab:prefixes}),
and train a Prompt Expansion model.
Having prepended the queries with prefixes during training,
at test-time we prepend the queries with query-specific prefixes to determine the format of the expanded prompts.
In Table \ref{tab:variations}, 
assuming the query type is known and the query-specific prefix is used during inference,
we find that \texttt{PE: Multi-Prefix}
maintains higher text-image alignment 
against other Prompt Expansion models.

Other than handling varying levels of image specification,
prefixes have also been used for indicating multi-step prompt expansion.
The use of queries sets a clear task boundary between different query sets/mixtures, 
such that the style of expanded prompt for one query does not interfere with others.
This also means we can continue to add new query types, their data, and prefixes over time, without risking interference with respect to other query types.
For a given dataset, if we assign a prefix for each data point,
the trade-off is that assigning (more) prefixes (further) dilutes the number of samples per prefix.
Encountering fewer samples per prefix may result in underfitting per prefix's data,
in contrast to fitting on the whole prefix-less dataset.
To balance this trade-off, 
the Prompt Expansion models in the main results
only make use of the \texttt{MSTP} prefix in the training dataset and during multi-step inference,
and
does not assign any other prefixes.
Models trained on other prefixes are used in \texttt{PE: Multi-Prefix} (query-specific) and \texttt{PE: Prefix Dropout} (query-agnostic) models in Table \ref{tab:variations}.
 \texttt{PE: Multi-Prefix} makes use of all augmentative mixtures, assigns prefixes to queries during training (though replacing \texttt{GRD} and  \texttt{SPCT} with \texttt{DTL}), and relies on the type-specific prefix during inference.

\begin{table*}[hb]
\caption{
Comparison of PE variations:
The base Prompt Expansion (PE) model maps varying query lengths to a fixed expanded prompt; we evaluate the ablation where each query maps to the subsequent slightly-longer query, resulting in varying output prompt length.
We also ablate the use of all prefixes and all mixtures in the dataset.
We sub-ablate this further by comparing the use of prefixes to perform query-specific inference (i.e. the model knows what kind of query it is handling) against
query-agnostic inference (through the use of Prefix Dropout).
}
\label{tab:variations}
    \begin{subtable}[h]{\textwidth}
        \centering
        \hspace*{-3cm}
        \resizebox{1.4\textwidth}{!}{%
        \begin{tabular}{llcccc}
        \hline \hline
         \multirow{2}{2.5cm}{\textbf{Method}} &  \multirow{2}{2cm}{\textbf{Prompt Category}}
          & \multirow{2}{4cm}{\textbf{Aesthetics \phantom{(MUSIQ-AVA)} (MUSIQ-AVA) $\uparrow$}} & \multirow{2}{4cm}{\textbf{Text-Image Alignment (COCA(q, I(p))) $\uparrow$}} 
          & \multirow{2}{4cm}{\textbf{Diversity ($\sigma_{p}$)) $\uparrow$}}
          & \multirow{2}{4cm}{\textbf{Output prompt length (mean+std)}}
          \\ \\  \hline

\multirow{7}{2.5cm}{Straight-Query Generation} & Task: Abstract 	& 5.114 $\pm$ 0.528 &	0.117 $\pm$ 0.0106 
& 0.00606 $\pm$ 0.00286 &	2.927 $\pm$  2.249 	\\  
 & Task: Concrete 	& 	5.141 $\pm$ 0.505 &	0.126 $\pm$ 0.0137

 &	0.00596 $\pm$ 0.00281 &	7.590 $\pm$  2.893 	\\  
 & Length: Short 	& 	5.0632 $\pm$ 0.516 &	0.117 $\pm$ 0.0106

 &	0.00611 $\pm$ 0.00289 &	2.196 $\pm$  0.686	\\  
 & Length: medium 	& 	5.142 $\pm$ 0.523 &	0.125 $\pm$ 0.0140

 &	0.00561 $\pm$ 0.00265 &	5.417 $\pm$  1.114 	\\  
 & Length: long 	& 	5.164 $\pm$ 0.513 &	0.134 $\pm$ 0.0180

 &	0.00571 $\pm$ 0.00268 &	11.525 $\pm$  5.773 	\\  
 & Dataset: WAA 	& 	5.128 $\pm$ 0.517 &	0.122 $\pm$ 0.0130

 &	0.00601 $\pm$ 0.00283 &	5.263 $\pm$  3.486	\\  
 & Dataset: PartiPrompts 	& 5.105 $\pm$ 0.523 &	0.134 $\pm$ 0.0189

 &	0.00535 $\pm$ 0.00253 &	9.156 $\pm$  7.483 	\\  \hline

\multirow{7}{2.5cm}{Few-shot Prompting} & Task: Abstract 	& 	 5.289 $\pm$ 0.560 & 0.102 $\pm$ 0.0161 
& 0.00780 $\pm$ 0.00372 & 11.218 $\pm$  3.0391 	\\  
 & Task: Concrete 	& 	5.293 $\pm$ 0.530 & 0.117 $\pm$ 0.0176 
 & 0.00726 $\pm$ 0.00346 & 17.0727 $\pm$  3.0695 	\\  
 & Length: Short 	& 	5.274 $\pm$ 0.568 & 0.100 $\pm$ 0.0164 
 & 0.00794 $\pm$ 0.00379 & 9.911 $\pm$  3.0202 	\\  
 & Length: medium 	& 	5.322 $\pm$ 0.522 & 0.117 $\pm$ 0.0181 
 & 0.00711 $\pm$ 0.00338 & 12.842 $\pm$  2.620 	\\  
 & Length: long 	& 	5.293 $\pm$ 0.552 & 0.127 $\pm$ 0.0198 
 & 0.00668 $\pm$ 0.00318 & 11.420 $\pm$  3.502 	\\  
 & Dataset: WAA 	& 	5.291 $\pm$ 0.545 & 0.109  $\pm$  0.0186 
 & 0.00753 $\pm$  0.00359 & 14.169  $\pm$  3.401 	\\  
 & Dataset: PartiPrompts 	& 	5.305 $\pm$  0.559 & 0.127 $\pm$  0.0233
 & 0.00658 $\pm$  0.00312 & 16.138 $\pm$   4.120 	\\  \hline
 
 \multirow{7}{2.5cm}{Prompt Expansion (PE)} & Task: Abstract 	& 	5.232 $\pm$ 0.596 & 0.112 $\pm$ 0.0118 
 & 0.00765 $\pm$ 0.00362 & 34.736 $\pm$  9.0216 	\\  
 & Task: Concrete 	& 	5.260 $\pm$ 0.567 & 0.120 $\pm$ 0.0154 
 & 0.00724 $\pm$ 0.00342 & 39.362 $\pm$  8.167 	\\  
 & Length: Short 	& 	5.233 $\pm$ 0.593 & 0.112 $\pm$ 0.0122 
 & 0.00780 $\pm$ 0.00369 & 34.287 $\pm$  8.850 	\\  
 & Length: medium 	& 5.227 $\pm$ 0.577 & 0.119 $\pm$ 0.0153 
 & 0.00707 $\pm$ 0.00334 & 37.165 $\pm$  8.169	\\  
 & Length: long 	& 5.216 $\pm$ 0.584 & 0.128 $\pm$ 0.0200 
 & 0.00669 $\pm$ 0.00316 & 42.645 $\pm$  9.129 	\\  
 & Dataset: WAA 	& 	5.246 $\pm$ 0.582 & 0.116 $\pm$ 0.0143 
 & 0.00744 $\pm$ 0.00352 & 37.054 $\pm$  8.909 	\\  
 & Dataset: PartiPrompts 	& 5.173 $\pm$ 0.590 & 0.129 $\pm$ 0.0208 
 & 0.00660 $\pm$ 0.00311 & 40.467 $\pm$  10.276 	\\  \hline

 \multirow{7}{2.5cm}{PE: \rft} 
 & Task: Abstract 	& 	 6.183 $\pm$  0.505 & 0.100 $\pm$  0.0162 
 & 0.00796 $\pm$  0.00380 & 42.0169 $\pm$  7.986 \\ 
 & Task: Concrete 	& 	6.183 $\pm$  0.451 & 0.115 $\pm$  0.0184 
 & 0.00744 $\pm$  0.00353 & 46.143 $\pm$  8.088 \\
 & Length: Short 	& 	6.172 $\pm$  0.510 & 0.0997 $\pm$  0.0167 
 & 0.00805 $\pm$  0.00384 & 41.985 $\pm$  7.978 \\ 
 & Length: medium 	& 6.210 $\pm$  0.443 & 0.115 $\pm$  0.0187 
 & 0.00729 $\pm$  0.00345 & 43.126 $\pm$  7.551 \\ 
 & Length: long 	& 6.176 $\pm$  0.461 & 0.125 $\pm$  0.0201
 & 0.00699 $\pm$  0.00332 & 49.49 $\pm$  8.861 \\ 
 & Dataset: WAA 	& 	6.183 $\pm$  0.479 & 0.108 $\pm$  0.0189 
 & 0.00770 $\pm$  0.00366 & 44.0845 $\pm$  8.298 \\ 
 & Dataset: PartiPrompts 	& 6.189 $\pm$  0.464 & 0.125 $\pm$  0.0227 
 & 0.00686 $\pm$  0.00325 & 47.0 $\pm$  9.754 \\ 
  \hline \hline

\multirow{7}{2.5cm}{PE: step-by-step expansion} & Task: Abstract 	& 	5.191 $\pm$ 0.539 & 0.114 $\pm$ 0.0120 
& 0.00756 $\pm$ 0.00360 & 11.161 $\pm$  3.941 	\\  
 & Task: Concrete 	& 	5.179 $\pm$ 0.505 & 0.124 $\pm$ 0.0141 
 & 0.00670 $\pm$ 0.00318 & 19.344 $\pm$  6.125 	\\  
 & Length: Short 	& 	5.173 $\pm$ 0.520 & 0.114 $\pm$ 0.0123 
 & 0.00765 $\pm$ 0.00365 & 9.888 $\pm$  2.719	\\  
 & Length: medium 	& 	5.187 $\pm$ 0.527 & 0.122 $\pm$ 0.0149 
 & 0.00663 $\pm$ 0.00316 & 15.405 $\pm$  3.787 	\\  
 & Length: long 	& 	5.161 $\pm$ 0.526 & 0.131 $\pm$ 0.0184 
 & 0.00617 $\pm$ 0.00290 & 25.926 $\pm$  9.401 	\\  
 & Dataset: WAA 	& 	5.185 $\pm$ 0.522 & 0.119 $\pm$ 0.0139 
 & 0.00713 $\pm$ 0.00339 & 15.261 $\pm$  6.579 	\\  
 & Dataset: PartiPrompts 	& 	5.144 $\pm$ 0.529 & 0.132 $\pm$ 0.0198 
 & 0.00610 $\pm$ 0.00289 & 21.582 $\pm$  12.526 	\\  \hline

\multirow{7}{2.5cm}{PE: Multi-Prefix} & Task: Abstract 	& 	 5.653 $\pm$ 0.610 & 0.116 $\pm$ 0.0153 
& 0.00658 $\pm$ 0.00315 & 20.198 $\pm$  8.617 	\\  
 & Task: Concrete 	& 5.718 $\pm$ 0.617 & 0.123 $\pm$ 0.0162
 & 0.00630 $\pm$ 0.00301 & 21.569 $\pm$  7.280	\\  
 & Length: Short 	& 5.663 $\pm$ 0.619 & 0.116 $\pm$ 0.0156 
 & 0.00652 $\pm$ 0.00312 & 20.196 $\pm$  8.0729
 	\\  
 & Length: medium 	& 5.744 $\pm$ 0.603 & 0.125 $\pm$ 0.0165 
 & 0.00627 $\pm$ 0.00298 & 22.471 $\pm$  7.849 	\\  
 & Length: long 	& 	5.709 $\pm$ 0.639 & 0.126 $\pm$ 0.0163 
 & 0.00616 $\pm$ 0.00293 & 21.976 $\pm$  6.117	\\  
 & Dataset: WAA 	& 5.705 $\pm$ 0.609 & 0.122 $\pm$ 0.0156 
 & 0.00638 $\pm$ 0.00305 & 20.994 $\pm$  6.833 	\\  
 & Dataset: PartiPrompts 	& 	5.729 $\pm$ 0.635 & 0.131 $\pm$ 0.0161 
 & 0.00588 $\pm$ 0.00278 & 23.692 $\pm$  6.501 	\\  \hline

\multirow{7}{2.5cm}{PE: Prefix Dropout} & Task: Abstract 	& 	 5.394 $\pm$ 0.618 & 0.116 $\pm$ 0.0147 
& 0.00667 $\pm$ 0.00320 & 18.217 $\pm$  8.298 	\\  
 & Task: Concrete 	& 	5.457 $\pm$ 0.631 & 0.120 $\pm$ 0.0149
 & 0.00640 $\pm$ 0.00308 & 19.264 $\pm$  8.672 	\\  
 & Length: Short 	& 5.384 $\pm$ 0.637 & 0.117 $\pm$ 0.0152 
 & 0.00660 $\pm$ 0.00317 & 18.377 $\pm$  8.472 	\\  
 & Length: medium 	& 5.401 $\pm$ 0.619 & 0.122 $\pm$ 0.0167
 & 0.00630 $\pm$ 0.00302 & 20.212 $\pm$  8.729 	\\  
 & Length: long 	& 	5.445 $\pm$ 0.608 & 0.122 $\pm$ 0.0158 
 & 0.00624 $\pm$ 0.00300 & 19.903 $\pm$  7.868 	\\  
 & Dataset: WAA 	& 	5.425 $\pm$ 0.625 & 0.118 $\pm$ 0.0149
 & 0.00654 $\pm$ 0.00314 & 18.741 $\pm$  8.504 	\\  
 & Dataset: PartiPrompts 	& 5.374 $\pm$ 0.614 & 0.129 $\pm$ 0.0174 
 & 0.00583 $\pm$ 0.00280 & 22.148 $\pm$  7.359 	\\  \hline \hline
        \end{tabular}%
        }
    \end{subtable}
\end{table*}
\clearpage

\subsection
{Query Ambiguity with \texttt{Prefix Dropout}}

At test-time, we may not be able to assume we know the query type or which prefix to use.
Thus, it is necessary for the Prompt Expansion model to 
infer the query type such that it can return appropriate expanded prompts (e.g. semantic vs detailed expansion).
The Prompt Expansion model will be required to (implicitly) infer the query type correctly, then perform type-specific inference. 
To tackle this, we develop \pd{}, a curriculum learning approach where we assign a prefix to each query type mixture, and train the Prompt Expansion model on the
data with prefixes, but with a linearly increasing dropout rate (starting from 0.4 dropout rate to gradually 1.0 dropout rate), where the prefixes are randomly dropped out. 
This allows the model to develop the representations needed to perform the task,
and then gradually reassign weights
whose activations would trigger these representations. 
The model learns during training that the prefixes are not a reliable indicator. 
We observed during training that the distribution of sampled expanded prompts would initially be either one query type or the other for a given query (implicit classification of the query type, i.e. type inference), but as the model begins to encounter a higher dropout rate such that no prefixes are available, it begins to diversify the generated samples such that there is a mix of sentences from all query types. 
During training, \texttt{PE: Prefix Dropout} makes use of all the augmentative mixtures, uses all the original prefixes, and during inference does not require the prefix.
As seen in Table \ref{tab:variations}'s results for \texttt{PE: Prefix Dropout},
\pd{} can resolve query type ambiguity at test-time with output samples from multiple types, while also retaining text-image (query-prompt) alignment.

\begin{figure*}[ht]
    \centering
    \includegraphics[width=.3\columnwidth]{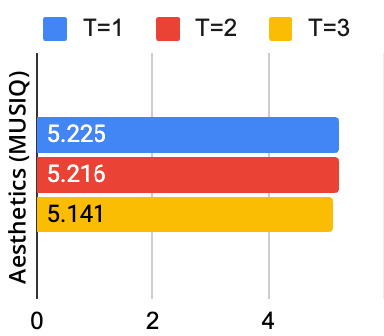}
    \includegraphics[width=.3\columnwidth]{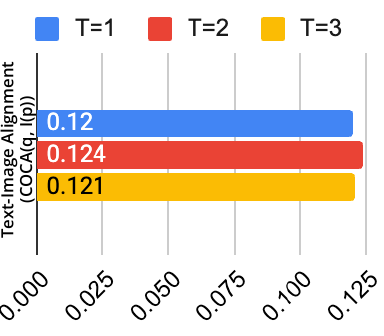}
    \includegraphics[width=.3\columnwidth]{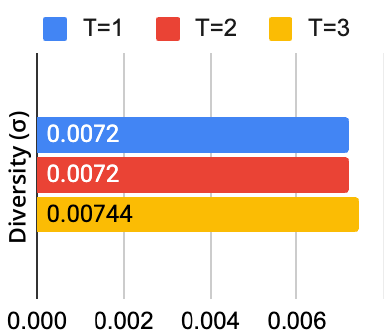}
    \caption{Evaluating consistency in Multi-Step Prompt Expansion:
            We find that the diversity of Prompt Expansion is consistent over number of steps.
            }
    \label{fig:num_expansion_steps}
\end{figure*}
\begin{figure*}[ht]
    \centering
    \includegraphics[width=.3\columnwidth]{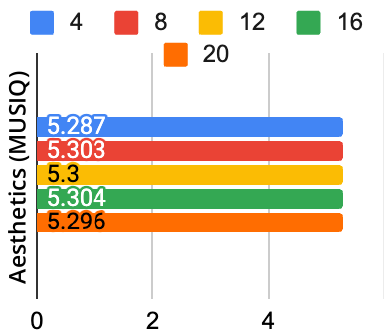}
    \includegraphics[width=.3\columnwidth]{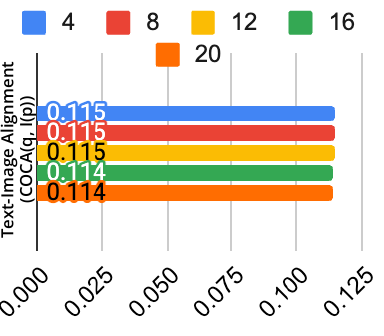}
    \includegraphics[width=.3\columnwidth]{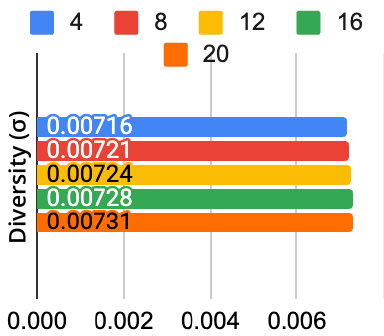}
    \caption{Evaluating consistency with scale: 
            We find that the diversity of Prompt Expansion is consistent over 
            number of expanded prompts per step.
            }
    \label{fig:num_expansion_prompts}
\end{figure*}

\section{Multi-Step Prompt Expansion}
\label{sec:multistep}

Rather than continuously increasing expanded prompt sequence length, 
we construct a set/mixture that maps an expanded prompt from step $t-1$ to an expanded prompt for step $t$.
Using a re-fine-tuned model, we generate expanded prompts on held-out queries, and iteratively generate prompts upon the previous step's prompts.
We re-fine-tune the model on this multistep dataset. 

As demonstrated in Figure \ref{fig:flow},
few-shot prompting of prompts to queries return multiple steps of expansion based on input length and output length. This enables multi-step prompt expansion for shorter queries. In the initial model, there is a maximum observed output length in our training dataset, and thus an expanded prompt may tend to stop expanding in length at this limit. Additionally, the text-to-text model also has a token length limit. 

To enable expanded prompts past such limits, 
we first construct a paired training set/mixture of expanded prompts to next-step expanded prompts.
The expanded prompt passed to the Prompt Expansion model returns expanded prompts (of similar output length but replaced details).
We reuse the Prompt Expansion model by passing the expanded prompt and receiving a set of variant expanded prompts of alternative details. We then append sentences with additional details to the input prompt.
We repeat this until the input prompt length reaches the token limit.
In practice, if the query exceeds the token limit, it will need to be truncated.

We re-fine-tune the Prompt Expansion model with the multi-step data, 
and add the \texttt{MSTP} prefix.
As the user is reusing the previous step's prompt, we know that the query type is that of "multi-step", 
and know from Table \ref{tab:variations} (\texttt{PE: Multi-Prefix}) that a prefix can return better metric performance if the query type is known and its prefix is used.
We also may wish to avoid the performance drop that may arise from having output sequences of varying lengths,
as seen in Table \ref{tab:variations} (\texttt{PE: step-by-step expansion}).
\texttt{PE: step-by-step expansion} is a supporting ablation that mapping queries to the full expanded prompt (as opposed to the subsequent query) is not ideal.

After generating $N=4$ expanded prompts from the initial query at step $t=0$, at step $t=1$ we then subsequently generate $N$ expanded prompt for each expanded prompt, and so on, resulting in $N^{t+1}$ expanded prompts in a set to evaluate.
In Figure \ref{fig:num_expansion_steps}, we observe
consistent diversity across number of expansion steps.
We evaluate against $N^{t+1}$ expanded prompts at the $t$-th step.
We observe that diversity is consistent in a multi-step setting over several iterations, where the expanded prompt from a previous iteration is passed back into the model to return another set of expanded prompts. 

While Figure \ref{fig:num_expansion_steps} evaluates the consistency of Prompt Expansion over the number of steps, 
Figure \ref{fig:num_expansion_prompts} evaluates its consistency over the number of expanded prompts generated.
We observe that the diversity of the generated images continue to increase,
and aesthetics and text-image alignment remains consistent.

\newpage
\begin{table*}[t]
\vspace{-1cm}
\centering
	\begin{minipage}{.75\columnwidth}
		\centering
    \resizebox{\textwidth}{!}{%
        \begin{tabular}{ll|cc}
        \hline \hline
        & 
        \textbf{flavor}
          & \textbf{COCA(q, I(p)) $\uparrow$} & \textbf{COCA(p, I(p))) $\uparrow$} \\ \hline
Top-5 flavors 
& art deco		 	& 0.163	& 0.169 \\
& vorticism		 	& 0.162	& 0.168\\
& classical realism		 	& 0.161	& 0.165\\
& fine art		 	& 0.155	& 0.166\\
& figurative art		 	& 0.159	& 0.161\\ \hline \hline
Worst-5 flavors
& academic art		 	& 	0.096& 0.110\\
& international gothic		 	& 0.107	& 0.098\\
& pixel art		 	& 0.085	& 0.114\\
& arte povera		 	& 0.093	& 0.102\\
& generative art		 	& 0.102	& 0.088\\ \hline \hline
        \end{tabular}%
        }
	\end{minipage}
    \begin{minipage}{\columnwidth}
		\centering
		\includegraphics[width=\columnwidth]{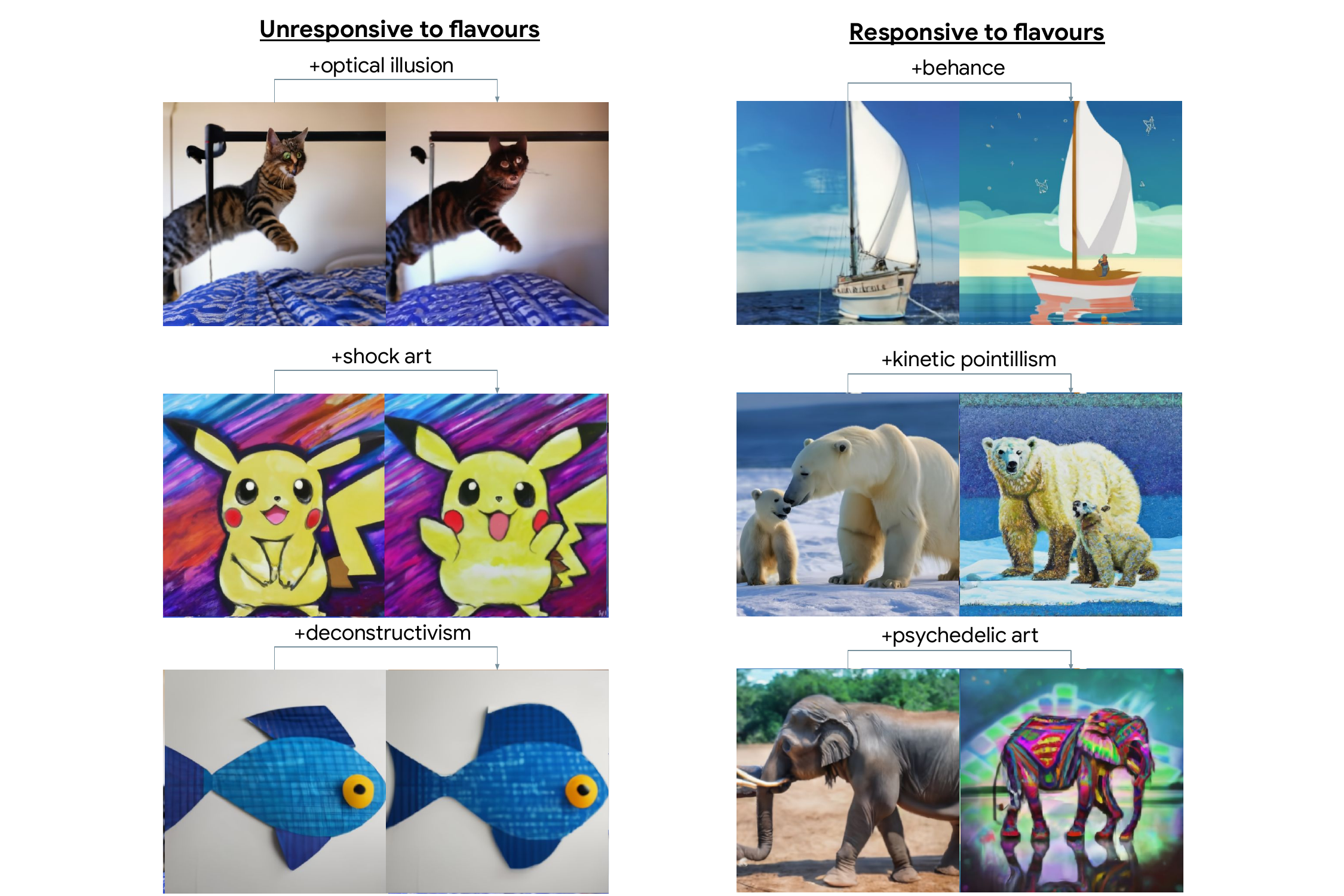}
    \end{minipage}
	\caption{
	\textbf{Top:} 
	The average query-image embedding distance and prompt-image embedding distance for flavors are ranked.
	\textbf{Bottom:}
	For the same seed, 
	image generation does not respond to the insertion to some flavors, while responding to others.
	}
	\label{tab:flavor_probe_full}
\end{table*}

\section{Probing the alignment between text-to-text and text-to-image models}
\label{sec:probe}

Using the base model, we probe the ability of the Prompt Expansion model in rendering aesthetically-pleasing and diverse images.
We observe that the text-to-image model has difficulty in rendering certain prompts (e.g. human hands, faces).
This can become an issue when the Prompt Expansion model generates prompts such that the text-to-image model cannot render its image.

We begin by evaluating how well flavors can be rendered.
Compared to specific details in captions, flavors have a higher frequency of overlap / re-use between different expanded prompts.
As such, we would be able to evaluate how well a flavor is rendered across multiple expanded prompts.
We evaluate the COCA similarity between the query text embedding and generated image embedding, 
as well as the COCA similarity between the prompt text embeddings and generated image embedding.

As shown in Table \ref{tab:flavor_probe_full} (Top), sorted by the average of the two COCA scores, 
the similarity scores between the Top-5 highest flavors diverges greatly from that of the Worst-5 flavors.
Referring to Table \ref{tab:flavor_probe_full} (Bottom),
where we generate two images (without and with flavor) with the same seed,
we can visually observe that flavors in the left column are unresponsive (e.g. the seeded image does not change much), while the flavors in the right column visually change the image.
Probing the difficulty in rendering certain prompts
motivates the need to re-fine-tune the Prompt Expansion model to align it with the target text-to-image model (Section \ref{sec:pemodel}).

\clearpage
\section{Prompt Tuning avoids a skewed token distribution}

\begin{figure*}[ht]
     \centering
     \caption{
    Examples of challenges mitigated in the current Prompt Expansion Model. 
    }
     \begin{subfigure}[t]{\textwidth}
         \centering
         \includegraphics[width=.45\textwidth]{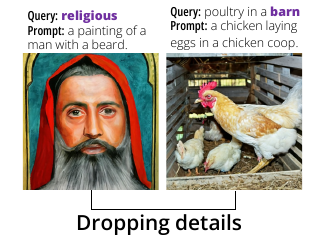}
         \caption{
         Dropping details from step-by-step expansion:
         Qualitatively supporting the findings from Table \ref{tab:variations}, training on variable output sequence length and mapping step-by-step (\texttt{PE: step-by-step expansion}) encounters challenges that others do not.
    We observe that a drop of details may occur (e.g. {\color{purple}barn} is not mentioned in the prompt for "poultry in a barn"), thus resulting in lower text-image alignment, especially for concrete / long queries.}
     \end{subfigure}
     \begin{subfigure}[t]{\textwidth}
         \centering
         \hspace*{-2cm}
         \includegraphics[width=1.3\textwidth]{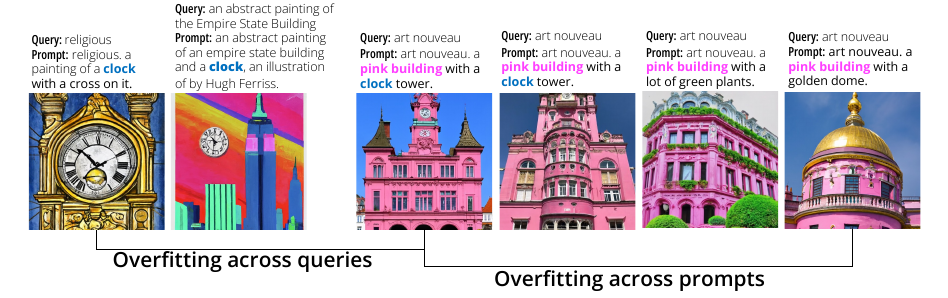}
         \caption{
         Overfitting from fine-tuning:
         Qualitatively supporting the findings from Table \ref{tab:model_config}, Prompt Tuning alleviates certain challenges that Fine Tuning (end-to-end) may present. 
    We also observe overfitting to the training distribution, where tokens (e.g. {\color{blue}clock}, {\color{magenta}pink building}) are repeated, regardless of same or different query. }
     \end{subfigure}
\label{fig:bad_examples}
\end{figure*}

With Table \ref{tab:model_config}, we establish the training setup of the PaLM 2 model.
We evaluate fine-tuning against prompt-tuning, and a smaller 1B against a larger 24B parameter model.
We show that prompt-tuning a 1B PaLM 2 model is the optimal setup, achieving pareto-optimality in parameter count, aesthetics, text-image alignment, and diversity scores compared to other configurations.
In-line with \citet{Welleck2020Neural}'s evaluation of repetition, we also evaluate the repetitiveness of expanded prompt generation.
We observe that fine-tuned models returned a higher repetition rate, indicating a potentially skewed token distribution, likely towards tokens seen during training.
This manifests in expanded prompts as recurring terms appearing regardless of varying queries, and even recurring terms for the same query (examples in Figure \ref{fig:bad_examples}b).

\section{Fixed/long length is the optimal prompt length compared to variable/short length}

In Table \ref{tab:variations}, 
we ablate the sequence length of the expected output prompt.
We do this by comparing two query-prompt mapping settings.
When we few-shot prompt a prompt $p$ to a (shorter) query $q_{n}$, where $n$ is the $n$th-order query, we then few-shot prompt the query $q_{n}$ to a (shorter) query $q_{n-1}$. We repeat this till query $q_{1}$ , until we have a set of $n$ queries mapped to each $p$.
While \texttt{Prompt Expansion} maps each $q_i \mapsto p$ for every $i \in n$,
\texttt{PE: step-by-step expansion} maps each (shorter) query $q_i$ to the next (longer) query $q_{i+1}$ such that
$q_i \mapsto q_{i+1}$.

Though the variance of output sequence length between \texttt{PE: step-by-step expansion} and \texttt{Prompt Expansion} is similar, the mean length is much higher for the latter than the former. 
We find that \texttt{PE: step-by-step expansion} consistently underperforms \texttt{Prompt Expansion} across varying query types across all metrics. 
Qualitatively we also find that the step-by-step expansion setup also results in drop of query details, as shown in
Figure \ref{fig:bad_examples}(left).

{
\begin{table*}[htbp]
\caption{
Evaluating different model configurations,
namely model size and parameter tuning method.
}
\label{tab:model_config}
\centering
    \begin{subtable}[h]{\textwidth}
    \tiny
        \centering
        \resizebox{\textwidth}{!}{%
        \begin{tabular}{l|cccc}
        \hline
        \hline
        \textbf{Method}
          & \textbf{Aesthetics (MUSIQ-AVA) $\uparrow$} & \textbf{Text-Image Alignment (COCA(q, I(p))) $\uparrow$} & \textbf{Diversity ($\sigma_{p}$)) $\uparrow$} & \textbf{Repetition Rate ($r$)) $\downarrow$} \\ \hline
        Fine-Tune, 1B
        & 5.282 $\pm$ 0.570	& 0.122 $\pm$ 0.0170 & 0.00621 $\pm$ 0.00297 & 0.0101\\  \hline
        Fine-Tune, 24B
        & 5.411 $\pm$ 0.622	& 0.1216 $\pm$ 0.0166	& 0.00657 $\pm$ 0.00312 & 0.112 \\  \hline
        \rowcolor{blue!10} 
        Prompt-Tune, 1B
        & 5.225 $\pm$ 0.585	& 0.120 $\pm$ 0.0175	& 0.00720 $\pm$ 0.00341 & 0.00285 \\  \hline
        Prompt-Tune, 24B
        & 5.335 $\pm$ 0.625	& 0.119 $\pm$ 0.0177	& 0.00702 $\pm$ 0.00331 & 0.00715 \\  \hline 
        \hline
        \end{tabular}%
        }
    \end{subtable}
\end{table*}
\begin{table*}[htbp]
\caption{
Ablations for diversity in Prompt Expansion and its subsequent text-to-image generation.
}
\label{tab:div_ablations}
    \begin{subtable}[h]{\textwidth}
        \centering
        \resizebox{\textwidth}{!}{%
        \begin{tabular}{l|c|ccc}
        \hline \hline
        \textbf{Method} & \textbf{Hyperparameters}
          & \textbf{Aesthetics (MUSIQ-AVA) $\uparrow$} & \textbf{Text-Image Alignment (COCA(q, I(p))) $\uparrow$} & \textbf{Diversity ($\sigma_{p}$)) $\uparrow$} \\ \hline
\multirow{1}{3cm}{Straight-Query Gen.}	
& \multirow{1}{3cm}{4 random seeds} &	5.121 $\pm$ 0.519 &	0.125 $\pm$ 0.0147 &	0.00582 $\pm$ 0.00275 \\
\hline
\multirow{3}{3cm}{Prompt Expansion}	
& \multirow{1}{3cm}{Temperature = 0.1} 	& 	5.267 $\pm$ 0.534 &	0.119 $\pm$ 0.0177 &	0.00702 $\pm$ 0.00331 \\
& \multirow{1}{3cm}{Temperature = 0.5} 	& 	5.283 $\pm$ 0.5266 &	0.116 $\pm$ 0.0215 &	0.00683 $\pm$ 0.00326 \\
& \multirow{1}{3cm}{Temperature = 1.0} 	& 	5.225 $\pm$ 0.585	&0.120 $\pm$ 0.0175 &	0.00720 $\pm$ 0.00341\\  \hline
\multirow{2}{3cm}{Prompt Expansion}
& \multirow{1}{3cm}{decoder=greedy} 	& 5.294 $\pm$ 0.538 &	0.115 $\pm$ 0.0213 &	0.0 $\pm$ 0.0 \\
& \multirow{1}{4cm}{decoder=beam search} 	& 5.0839 $\pm$ 0.550 &	0.118 $\pm$ 0.0211 &	0.00426 $\pm$ 0.00332 \\  \hline

\multirow{2}{3cm}{Prompt Expansion}
& \multirow{1}{4cm}{+ text post-hoc filtering} 	& 5.318 $\pm$ 0.542	& 0.114 $\pm$0.0216	& 0.00726 $\pm$0.00346 \\
& \multirow{1}{4cm}{+ image post-hoc filtering} 	& 5.316 $\pm$0.562	& 0.113 $\pm$0.0215	& 0.00739 $\pm$0.00350 \\  \hline\hline
        \end{tabular}%
        }
    \end{subtable}
\end{table*}
}

\section{Diversity ablations}

We further evaluate diversity in the output images compared to other
ablation methods in Table \ref{tab:div_ablations}.
The baseline is
Straight-Query Generation, 
which generates 4 images with 4 random seeds. 

\noindent\textbf{Decoding. }
Prompt Expansion uses temperature-based decoding, with default temperature being 1.0.
The first set of ablations is varying temperature between 0 to 1. 
A higher temperature tends to result in more diverse output for text models using temperature-based decoding.
The next set of ablations is changing the decoding method, specifically greedy decoding and beam search decoding (with beam size 4).
While beam search can return 4 expanded prompts per query, 
greedy decoding returns 1 expanded prompt per query. 
For the latter, this means that the embedding variance of 1 image generated per expanded prompt $\sigma_p$ is 0.0.

\noindent\textbf{Post-hoc filtering. }
Our final set of ablations is to maximize diversity after text/image generation.
Specifically, post-hoc filtering aims to generate a large number (N=20) of prompts/images, 
and filter out 4 that return the highest combinatorial diversity score.
For post-hoc text filtering,
we return 20 prompts through Prompt Expansion, 
compute the COCA text embeddings for each prompt, 
then enumerate through each combination of 4 prompts, 
and the selected 4 prompts is the combination that has the highest variance in text embeddings.
For post-hoc image filtering, 
we return 20 prompts through Prompt Expansion and generate 1 image per prompt, 
compute the COCA image embeddings for each prompt, 
then enumerate through each combination of 4 images, 
and the selected 4 images is the combination that has the highest variance in image embeddings.

\newpage
\section{Human Rater Task}
\label{sec:rater}

\begin{figure*}
    \vspace{-2.25cm}
     \centering
     \begin{subfigure}[t]{
     \textwidth
     }
         \centering
         \includegraphics[
         height=.5\textheight
         ]{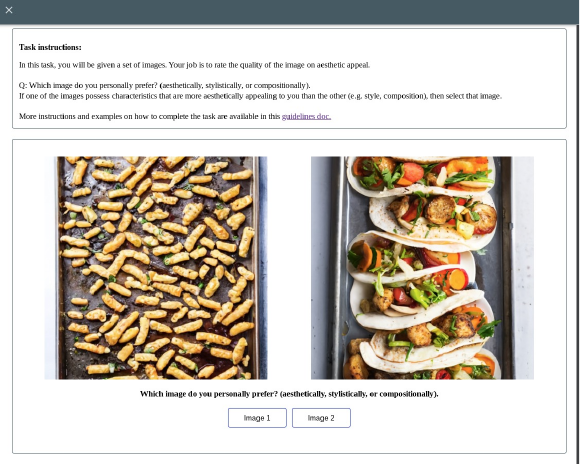}
         \caption{1x1.
    The question here is on aesthetics.
    No query is shown to the user.
    }
     \end{subfigure}
     \begin{subfigure}[t]{
     \textwidth
     }
         \centering
         \includegraphics[
         height=.5\textheight
         ]{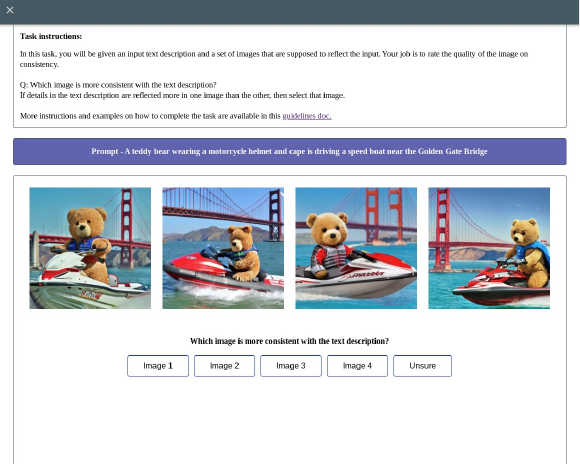}
         \caption{4x4.
        The question here is on text-image alignment.
        The query is shown to the user.
        }
     \end{subfigure}
\caption{
The user interface for our human rater experiments. 
}
\label{fig:raterui}
\end{figure*}

We retain the queries (n=700) and expanded prompts used in the automatic evaluation,
and evaluate them with human raters.
We perform a side-by-side (SxS) evaluation task, where we show a rater a set of images side by side for a given text query, and ask the rater to pick the best image.
We split the 700 queries into two sets of 350 for 1x1 and 4x4 settings.
Each instance was evaluated by 3 human raters, with the strong consensus (Table \ref{tab:consensus}) indicating consistency in preferences between raters.
Figure \ref{fig:raterui} shows the annotation interface for the rater experiment.

\noindent\textbf{Aesthetics.}
In this setting, raters are first provided instructions on how to evaluate Aesthetics. 
We describe it as follows:
\begin{quote}
In this task, please select the image that you \underline{personally prefer}. You may find this image more appealing due to aesthetic, stylistic or compositional qualities. Please only evaluate this image for aesthetic appeal, and ignore all other aspects (overall image quality).
\end{quote}
We provide examples of one image of high aesthetic appeal side-by-side with another image of low aesthetic appeal.
In addition to positive examples, we provide examples of images not to reject due to flaws in image quality. 
\begin{quote}
Please only evaluate this image for aesthetic appeal, and ignore all other aspects (e.g. overall image quality).
Image quality measures the visual flaws of the image. Examples of visual flaws include, watermarks, illegible text, hallucinated content, warped features or objects, blurriness, and out of place objects. If these flaws appear, this does not mean the image must be rejected. Only reject the image if the other image is one that you personally prefer (e.g. more appealing due to aesthetic, stylistic or compositional, but not image quality).
\end{quote}
Once the rater is acclimated with the metric, we proceed with the task and can ask them:
\begin{quote}
Q: Which image do you personally prefer? (aesthetically, stylistically, or compositonally).
\end{quote}

\begin{figure}[t]
    \centering
    \caption{
An overview of the 1x1 and NxN rater experiments.
A Query vs Prompt evaluation always takes place, 
but an NxN experiment requires a set of 4 images to be evaluated against each other before the 1x1 experiment.
}
    \includegraphics[width=\columnwidth]{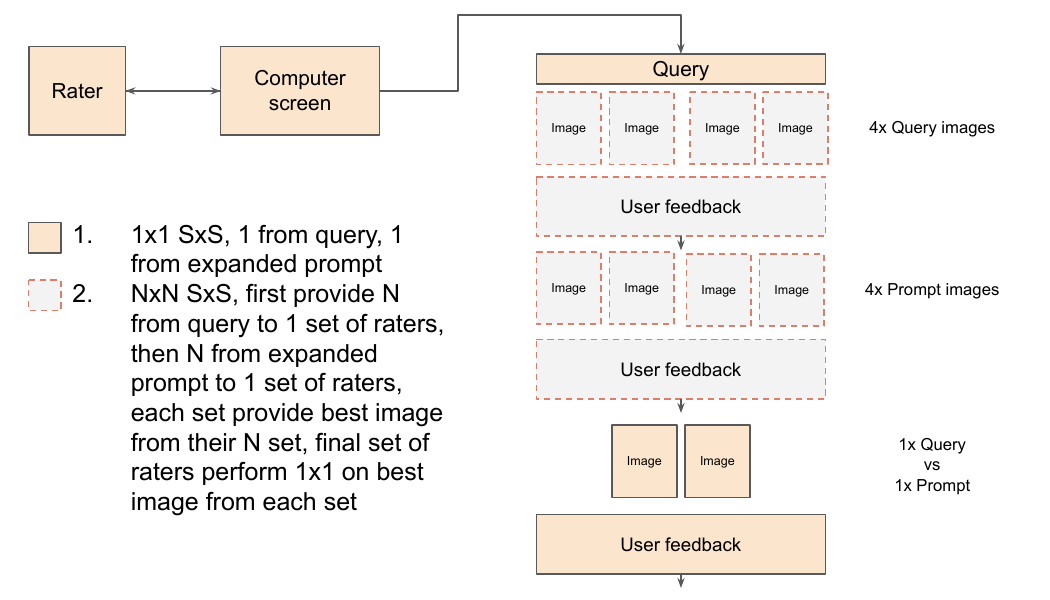}
    \label{fig:flow_raters}
\end{figure}

\noindent\textbf{Text-Image Alignment.}
We provide positive examples of images more aligned to the text description (query) compared to the other side-by-side image.
We similarly provide negative examples of images not pertaining to text-image alignment.
We describe the task to the rater as follows:
\begin{quote}
Please only evaluate this question for consistency, and ignore all other aspects of the image (image quality, or overall image preference).
\end{quote}
Furthermore, we add an option \texttt{Unsure}
when the rater perceives all images as equally likely to be picked.
Specifically, we describe to the raters to select this option as follows:
\begin{quote}
Select \texttt{Unsure} if:
\begin{itemize}
\item 
You do not understand the text prompt / it is unclear
\item 
Both images are equally consistent 
\item 
Both images are equally inconsistent 
\end{itemize}
\end{quote}

Once the rater is acclimated with the metric, we proceed with the task and can ask them:
\begin{quote}
Q: Which image is more consistent with the text description?
\end{quote}

\begin{table}[t]
\centering
\caption{
Consensus between Raters:
This details how many of the raters agree on an each Prompt-win image (i.e. number of raters that all picked the same Prompt-win image). 
}
\label{tab:consensus}
\begin{subtable}[h]{\columnwidth}
\centering
\tiny
\resizebox{\textwidth}{!}{%
\begin{tabular}{l|ccc}
\hline \hline
\textbf{Prompt-win Consensus} & \textbf{Consensus Agreement (3/3)} & \textbf{Consensus Agreement (2/3)} & \textbf{Consensus Agreement (0/3)} \\ \hline
1x1 Aesthetics & 0.450 $\pm$ 0.0103 &	0.549 $\pm$ 0.0103 &	0.0 $\pm$ 0.0  \\ \hline
4x4 Aesthetics &	0.396 $\pm$ 0.0167 &	0.604 $\pm$ 0.0167 &	0.0 $\pm$ 0.0 \\ \hline
1x1 Text-Image Alignment &	0.513 $\pm$ 0.146	& 0.446 $\pm$ 0.117 &	0.0407 $\pm$ 0.0288 \\ \hline
4x4 Text-Image Alignment	& 0.486 $\pm$ 0.150 &	0.481 $\pm$ 0.128 &	0.0333 $\pm$ 0.0259 \\ \hline
\hline
\end{tabular}%
}
\end{subtable}
\end{table}

\noindent\textbf{Random vs. Random (1x1).}
Can we attribute aesthetic preference of prompt expanded images over straight-query images to improved prompt construction / engineering? 
To answer this question, we perform a one-to-one side-by-side (1x1 SxS) comparison, 
where we pick the \textit{first} image generated from straight-query generation and the \textit{first} image generated from Prompt Expansion, then ask the raters to evaluate which one is better. 
This tests the ability of Prompt Expansion to reduce the need for manual prompt engineering by the user.
For both Aesthetics and Consistency, raters have 700 discrete rating decisions (we split to n=350, going through 
1,400
images).
The raters can pick between the options: Left, Right, (Unsure).
For the consistency task, we provide the Unsure option in case two images are equally good (or bad), since text-image alignment can be objectively evaluated. For aesthetics, we do not provide the Unsure option to make their aesthetic preference.

\noindent\textbf{Best vs. Best (4x4).}
Can we attribute aesthetic preference of prompt expanded images over straight-query images to increased diversity of samples? 
To answer this question, we perform a N-to-N side-by-side (NxN SxS) comparison followed by another 1x1 SxS comparison.
We begin by showing the rater N=4 images from straight-query generation.
We also show the rater N=4 images from Prompt Expansion.
The raters are not shown straight-query generation and Prompt Expansion images in separate stages; the 4x4 SxS is in one stage, and the rating decisions are shuffled.
The raters pick the best image out of N=4.
The raters can pick between the options: 1, 2, 3, 4, (Unsure).
Then the raters enter a second stage, 
where they are shown the best straight-query image against the best Prompt Expansion image for the same query, and asked to select the best. 
The raters can pick between the options: Left, Right, (Unsure).
For both Aesthetics and Consistency, raters have 2,100 discrete rating decisions (we split to a distinctly-different N=350, going through
7,000
images).

\newpage
\clearpage
\section{Datasets}
\label{sec:datasets}

Several datasets varying by modality, training-evaluation use, captioned/non-captioned, etc are used throughout the paper. We provide further details on all of them here.

\noindent
\textbf{Webli-Align \citep{jia2021scaling, chen2023pali}. }
\waa{}
is composed of two existing datasets Webli \citep{chen2023pali} and Align \citep{jia2021scaling}.
We filter out non-aesthetic images in Webli-Align with the MUSIQ \citep{ke2021musiq} aesthetics metric (refer to Sec \ref{sec:metrics}).
The filtered dataset contains 80k images.
We also perform a 70-20-10 train-val-test split. 
Within the train set, we split 50-50 for base and re-fine-tune training.

\noindent
\textbf{\crowdsource{}.}
\crowdsource{} is obtained through crowd-sourcing,
where we provide an interface for text-to-image generation (similar to \citet{gradio}), 
and users enter queries to generate images.
Users also have the option to upvote images that they liked, and we use this to filter for aesthetically-appealing images.
An example of an upvoted, aesthetically-appealing image is shown in Figure \ref{fig:flow}.
The filtered dataset contains 40k images.
We also perform a 70-20-10 train-val-test split. 
Within the train set, we split 50-50 for base and re-fine-tune training.

\noindent
\textbf{Base Prompt Expansion Dataset. }
The procedure for generating the base dataset is detailed in Section \ref{sec:pemdataset}.

\noindent
\textbf{Re-fine-tune Prompt Expansion Dataset. }
From the queries of the held-out set from the base dataset, 
we compute the expanded prompts with the base Prompt Expansion model (4 prompts per query).
We discard the original prompt mapped to the query, and replace it with the new expanded prompts.
For each expanded prompt, 
we compute the generated image.
From this probing set, we score the images in order to rank the images based on renderability with respect to its prompt text.
Our scoring metric is a weighted average of the query-image embedding distance and prompt-image embedding distance: 
$0.6 \times \texttt{COCA(q, I(p))} + 0.4 \times \texttt{COCA(p, I(p))}$.
We filter out query-prompt-image pairs that are below a threshold.
This filtered \texttt{\{query:prompt\}} dataset is thus used for re-fine-tuning.

While other scoring tools such as using MUSIQ scores or measuring embedding distance with respect to common image defects (e.g. distorted hands/faces) were also tested, 
we qualitatively find that this weighted average
as a general heuristic
can sufficiently filter out images that are non-renderable, unaligned to neither query nor prompt, and non-aesthetic. 
We also avoid completely filtering out every prompts/flavor that do not work. 
If the user intentionally requests for specific flavors / objects, 
we would need to generate expanded prompts that still fit the query.

\noindent
\textbf{(Evaluation) Webli-Align (WA) \citep{jia2021scaling, chen2023pali}. }
We sample n=500 queries from a test set of potential queries constructed by applying the PE dataset generation process of Section \ref{sec:pemdataset} (Image-to-Text Inversion + Query/Prompt Extraction) to \waa{} (WA) \citep{chen2023pali, jia2021scaling} images.
Queries are categorized 
based on whether they are 
abstract (WA=249),
concrete (WA=251), 
short (<4 words) (WA=224), 
medium length (4-7 words) (WA=134), 
or long length (>7 words) (WA=143). 

\noindent
\textbf{(Evaluation) PartiPrompts (PP) \citep{yu-parti}. }
We sample n=200 prompts from the PartiPrompts (PP) dataset of prompts that is designed to represent different domains and features of language (such as counting, negation, etc).
Queries are categorized 
based on whether they were 
short (<4 words) (PP=35), 
medium length (4-7 words) (PP=66), 
or long length (>7 words) (PP=66). 

\newpage
\section{COCA-Interrogator}
\label{sec:interog}

The \textit{Interrogator} \citep{clipin} approach to image-to-text inversion
requires an image-text model to probe what set of words and phrases are most similar to a given image, and concatenate the words/phrases together as the prompt text.
The prompt text computed through Interrogators are composed of (i) a caption, and (ii) a set of flavors.
The caption is a description of the content of the image (e.g. who, what, where, when).

We detail the implementation steps for COCA-Interrogator.
First, it requires an image captioning model (COCA fine-tuned on captioning \citep{yu2022coca}) to compute the image captions. 
Then, it computes flavors pertaining to the image, and appends them to the caption. 
The distance between the text embedding of the flavor and image embedding of the image is used to measure how likely the image manifests the given flavor. 
To curate the lists of flavors,
we first aggregate a large number of prompts written by users to generate images for text-to-image models. 
We split the prompts into words and phrases, measure its COCA distance w.r.t. the corresponding generated image, 
and filter out infrequent flavors.
Flavors are categorized by 
art form (e.g. vector art), 
artist (e.g. Maurycy Gottlieb),
medium (e.g. reddit contest winner),
style (e.g. neo-primitivism, photorealistic),
and other commonly-used phrases to prompt engineer images (e.g. monster manula, glamor pose, matte finish).
From each potential flavor, we compute the text embeddings, and measure the cosine similarity with respect to the image embeddings of the image. 
Both text and image embeddings are computed with COCA \citep{yu2022coca}.
We enumerate through each flavor across the categories 
to identify the set of flavors that maximize similarity. 
Flavors are organized by categories, and we ensure that at least one flavor from each
category is included.
Though COCA-Interrogator is used here, other image-to-text inversion methods can be used in-place,
such as CLIP-Interrogator \citep{clipin} or PEZ Dispenser \citep{wen2023hard}. 
We pursue the interrogator approach to inversion, given the high interpretability /
human readability of the output, and because it enables us to explicitly include
prompt-engineering flavors that govern the style and tone of the generated image.

\newpage
\begin{figure*}
\centering
\vspace*{-3cm}
\captionsetup{width=1.5\textwidth}
\hspace*{-3.5cm}
\begin{subfigure}[b]{1.5\textwidth}
\centering
\includegraphics[width=\textwidth]{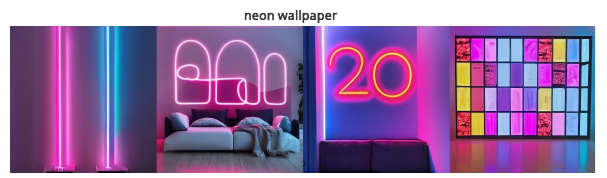}
\caption{Straight-Query Generation}
\end{subfigure}
\hspace*{-3.5cm}
\begin{subfigure}[b]{1.5\textwidth}
\centering
\includegraphics[width=\textwidth]{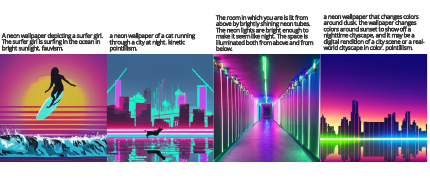}
\vspace*{-1cm}
\caption{Prompt Expansion}
\end{subfigure}
\hspace*{-3.5cm}
\begin{subfigure}[b]{1.5\textwidth}
\centering
\includegraphics[width=\textwidth]{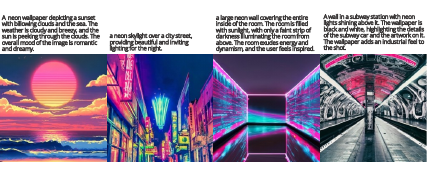}
\vspace*{-1cm}
\caption{Prompt Expansion (\rft)}
\end{subfigure}
\caption{
We compare images generated between Prompt Expansion and PE (\rft). 
Both explore different interpretations of the query, e.g. poster, city environments, rooms.
The re-fine-tuned model explores more types of locations and types of \textit{neon}.
}
\label{fig:demo13}
\end{figure*}

\begin{figure*}[t]
\centering
\vspace*{-3cm}
\captionsetup{width=1.5\textwidth}
\hspace*{-1.5cm}
\includegraphics[width=1.25\textwidth]{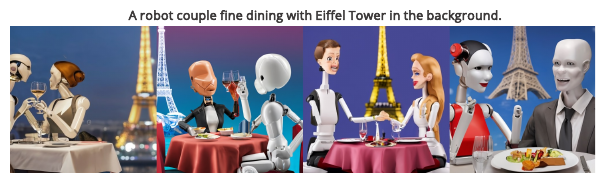}
\hspace*{-1.5cm}
\includegraphics[width=1.25\textwidth]{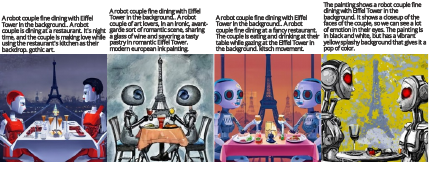}
\vspace*{-1cm}
\caption{
\textbf{Concrete / Long queries}:
Using a query from PartiPrompts,
we find both retainment and faithfulness to the original query, as well as stylistic variations in the subjects and environment.
}
\label{fig:demo11}
\end{figure*}

\begin{figure*}[t]
\centering
\vspace*{-0.45cm}
\captionsetup{width=1.5\textwidth}
\hspace*{-1.5cm}
\includegraphics[width=1.25\textwidth]{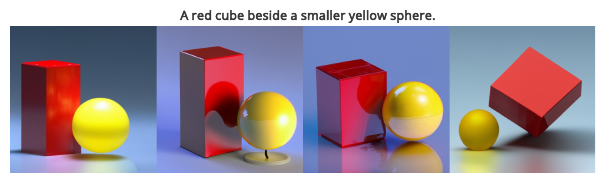}
\hspace*{-1.5cm}
\includegraphics[width=1.25\textwidth]{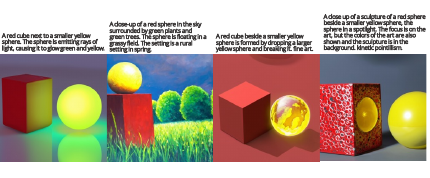}
\vspace*{-1cm}
\caption{
\textbf{Concrete queries}:
Using another query from PartiPrompts,
we find both retainment and faithfulness to the original query (e.g. size and colors of the objects), as well as  variations in the characteristics of the objects and environment.
}
\label{fig:demo12}
\end{figure*}

\begin{figure*}[t]
\centering
\vspace*{-3cm}
\captionsetup{width=1.5\textwidth}
\hspace*{-1.5cm}
\includegraphics[width=1.25\textwidth]{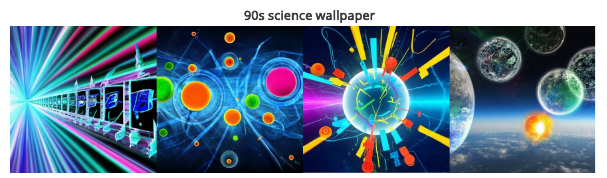}
\hspace*{-1.5cm}
\includegraphics[width=1.25\textwidth]{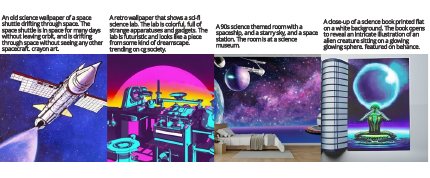}
\vspace*{-1cm}
\caption{
\textbf{Navigational / ideation queries}:
The interpretation of \textit{wallpaper} can vary by user intent. As a result, Prompt Expansion returns images across different mediums (e.g. poster wallpapers, room wallpapers). It also shows a variation of aesthetic styles referencing the 90s.
}
\label{fig:demo10}
\end{figure*}

\begin{figure*}[t]
    \centering
    \vspace*{-0.75cm}
    \captionsetup{width=1.5\textwidth}
    \hspace*{-2cm}
    \includegraphics[width=1.25\textwidth]{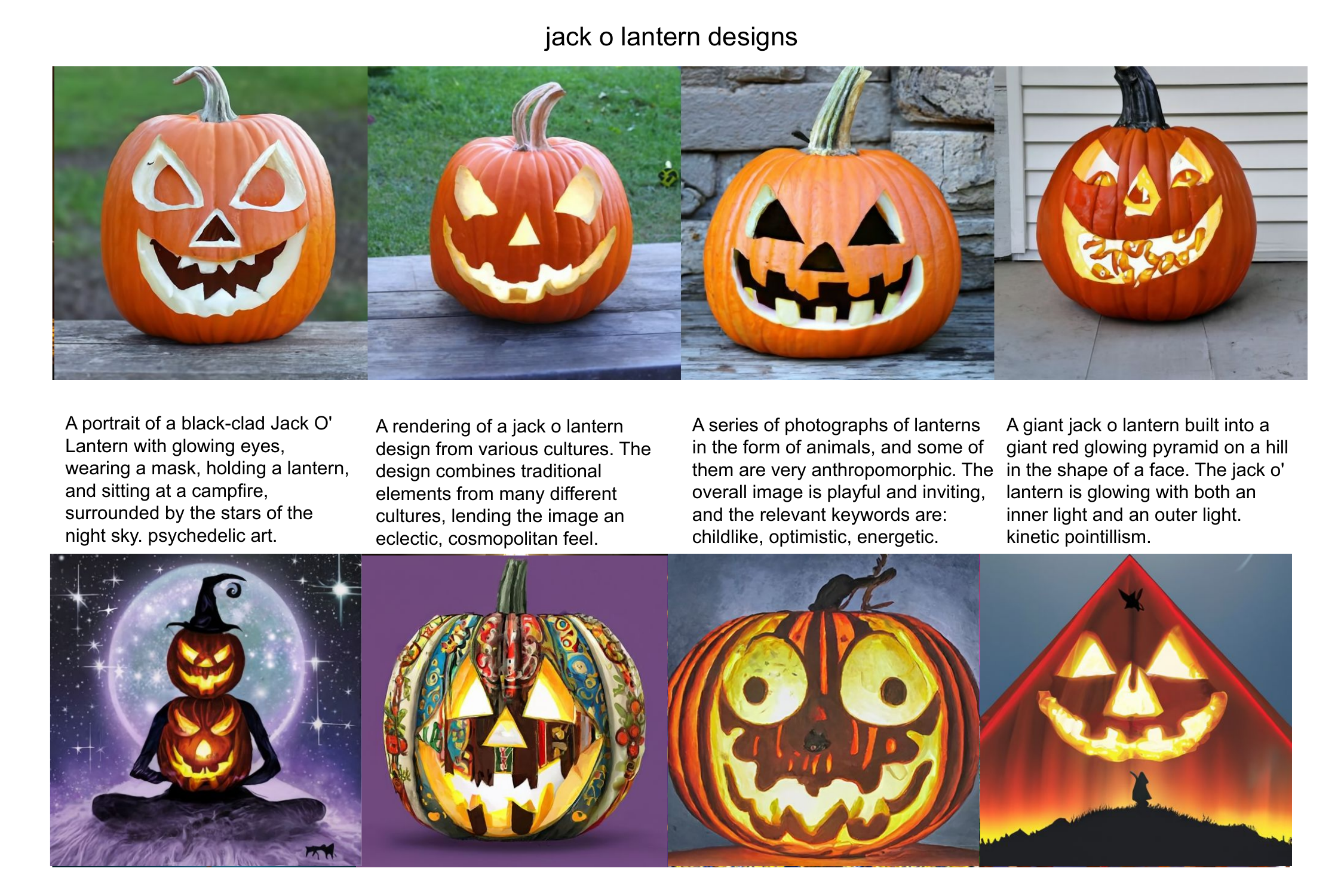}
    \vspace*{-0.75cm}
    \caption{
    \textbf{Navigational / ideation queries}:
    These types of abstract queries are intended to be exploratory, and the user does not have a specific image in mind necessarily. They are looking for "designs", so showing as many variations would improve the user experience.
    }
    \label{fig:demo4}
\end{figure*}

\clearpage

\end{document}